\documentclass[sn-pdflatex, sn-mathphys-num]{sn-jnl}
\usepackage{graphicx}%
\usepackage{multirow}%
\usepackage{amsmath,amssymb,amsfonts}%
\usepackage{amsthm}%
\usepackage{mathrsfs}%
\usepackage[title]{appendix}%
\usepackage{xcolor}%
\usepackage{textcomp}%
\usepackage{booktabs}%
\usepackage{natbib}
\usepackage{url}
\usepackage{listings}%
\usepackage{color}
\usepackage{makecell}
\usepackage{booktabs}
\usepackage{float}
\usepackage{subfigure}
\usepackage{subcaption}
\usepackage{hyperref}
\usepackage{placeins}

\newcommand{\bT}{{\bf T}}

\newcommand{\bY}{{\bf Y}}
\newcommand{\bz}{{\bf z}}

\newcommand{\bX}{{\bf X}}

\newcommand{\bW}{{\bf W}}

\newcommand{\bc}{\begin{center}}
\newcommand{\ec}{\end{center}}
\newcommand{\be}{\begin{equation}}
\newcommand{\ee}{\end{equation}}
\newcommand{\ba}{\begin{array}}
\newcommand{\ea}{\end{array}}
\newcommand{\bean}{\begin{eqnarray*}}
\newcommand{\eean}{\end{eqnarray*}}
\newcommand{\bea}{\begin{eqnarray}}
\newcommand{\eea}{\end{eqnarray}}

\newcommand{\ben}{\begin{enumerate}}
\newcommand{\een}{\end{enumerate}}
\newcommand{\bed}{\begin{itemize}}
\newcommand{\eed}{\end{itemize}}

\begin{document}

\title[Article Title]{Unicorn: U-Net for Sea Ice Forecasting with Convolutional Neural Ordinary Differential Equations}


\author[1]{\fnm{Jaesung} \sur{Park}}\email{wkdso0804@uos.ac.kr}
\equalcont{These authors contributed equally to this work.}

\author[2]{\fnm{Sungchul} \sur{Hong}}\email{shong@uos.ac.kr}
\equalcont{These authors contributed equally to this work.}

\author[1]{\fnm{Yoonseo} \sur{Cho}}\email{friend0429@uos.ac.kr}

\author*[1,2]{\fnm{Jong-June} \sur{Jeon}}\email{jj.jeon@uos.ac.kr}

\affil[1]{\orgdiv{Department of Statistical Data Science}, \orgname{University of Seoul}, \orgaddress{\street{163 Seoulsiripdaero}, \city{Seoul}, \postcode{02504}, \country{South Korea}}}
\affil[2]{\orgdiv{Department of Statistics}, \orgname{University of Seoul}, \orgaddress{\street{163 Seoulsiripdaero}, \city{Seoul}, \postcode{02504}, \country{South Korea}}}


\abstract{Sea ice at the North Pole is vital to global climate dynamics. However, accurately forecasting sea ice poses a significant challenge due to the intricate interaction among multiple variables. Leveraging the capability to integrate multiple inputs and powerful performances seamlessly, many studies have turned to neural networks for sea ice forecasting. This paper introduces a novel deep architecture named Unicorn, designed to forecast weekly sea ice. Our model integrates multiple time series images within its architecture to enhance its forecasting performance.
Moreover, we incorporate a bottleneck layer within the U-Net architecture, serving as neural ordinary differential equations with convolutional operations, to capture the spatiotemporal dynamics of latent variables. Through real data analysis with datasets spanning from 1998 to 2021, our proposed model demonstrates significant improvements over state-of-the-art models in the sea ice concentration forecasting task.  It achieves an average MAE improvement of 12\% compared to benchmark models. Additionally, our method outperforms existing approaches in sea ice extent forecasting, achieving a classification performance improvement of approximately 18\%. These experimental results show the superiority of our proposed model.} 

\keywords{Sea ice forecasting, Spatiotemporal forecasting, U-Net, Neural ODE}

\maketitle

\section{Introduction}\label{sec1}
Many studies have highlighted the close relationship between global climate change and the Arctic over the years, particularly emphasizing the role of sea ice \cite{Notz2016ObservedAS}. 
The phenomenon of Arctic amplification, where temperatures in the Arctic are rising at a rate four times higher than the global average, underscores the Arctic's position as an indicator of climate change \cite{Wunderling2020GlobalWD, Rantanen2022TheAH}. The diminishing of sea ice has led to increased surface temperatures and lower albedo, a proportion of radiation reflected by sea ice, setting off a chain of positive climate feedback mechanisms that further fuel Arctic amplification \cite{Screen2010TheCR, Pithan2014ArcticAD}.

\begin{figure}[t]
        \centering
        \includegraphics[width=0.7\linewidth]{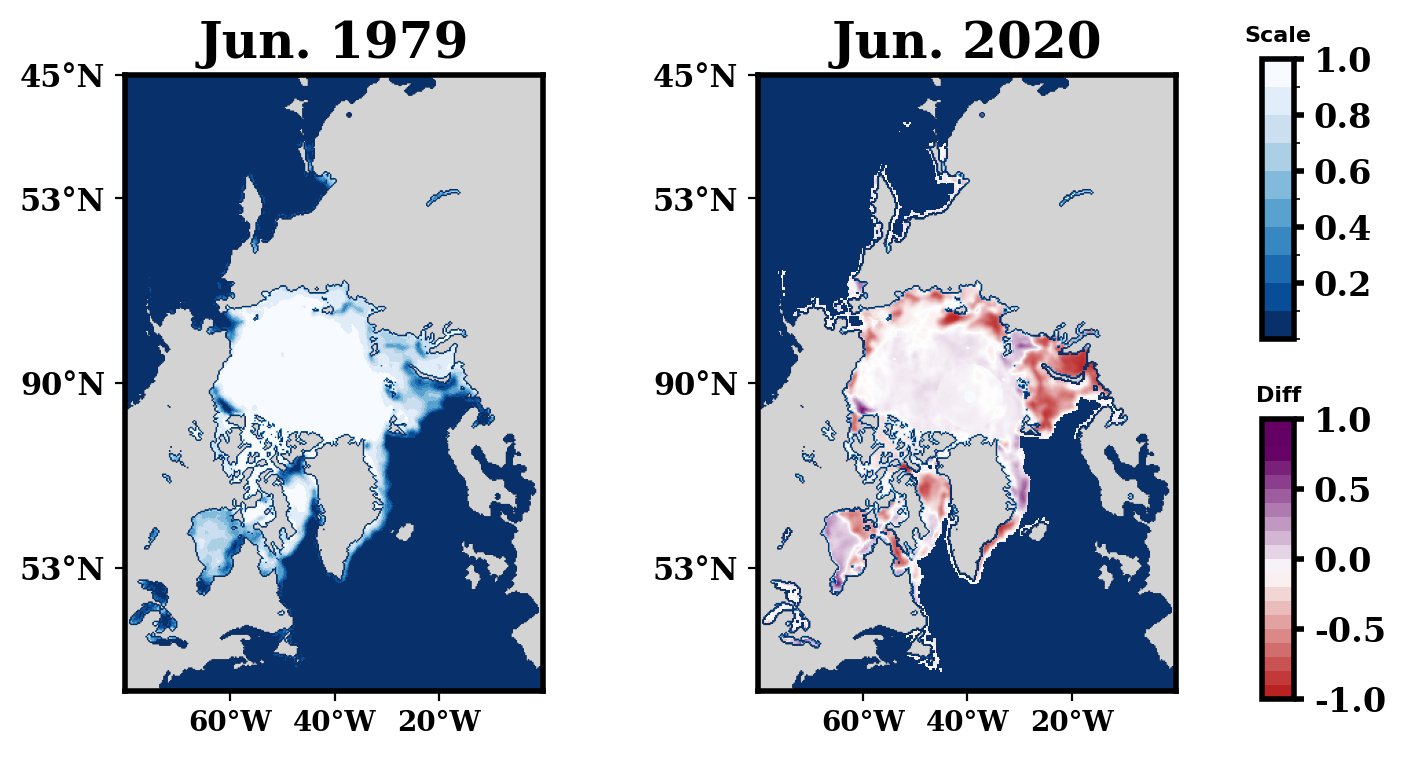}
        \caption{Monthly average sea ice concentration in June 1979 (left) and June 2020 (right). The colored regions highlight the difference in sea ice concentration between the two years.}
        \label{fig:intro}
\end{figure}

Prediction of sea ice loss is crucial as it affects the local Arctic ecosystem and weather conditions in the Northern Hemisphere \cite{Li2009SmallestAT, Kraemer2024AMT}. From the late 1970s to the mid-2000s, the Arctic experienced a rapid decrease in sea ice, with summer ice declining at rates up to 13\% per decade \citet{Wei2022PredictionOP}.
As shown in Figure \ref{fig:intro}, there is a clear decrease in June sea ice concentration between 1979 and 2020. With ongoing anthropogenic influence, this trend, coupled with the sharp reduction in multi-year ice and sea ice thickness, is expected to continue indicating both a shrinking and thinning of the Arctic sea ice cover \cite{Kwok2018ArcticSI, Sumata2023RegimeSI}. 
Projections based on the current pace of temperature rise suggest that we will experience a sea-ice-free Arctic summer by 2050 \cite{Notz2020ArcticSI, Kim2023ObservationallyconstrainedPO}.

\begin{figure}[t]
        \centering
        \includegraphics[width=0.7\linewidth]{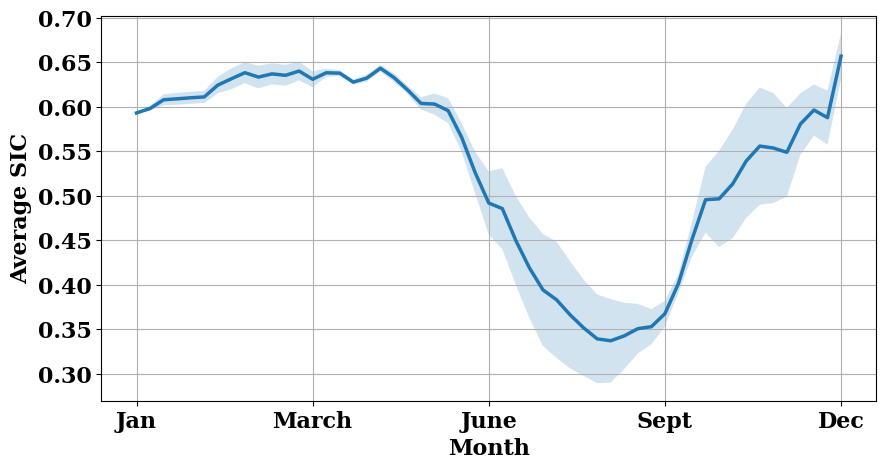}
        \caption{Trend of the averaged SIC spanning from 1998 to 2021. The blue band indicates the average $\pm 1$ standard deviation.}
        \label{fig:trend}
\end{figure}

\begin{figure}[t]
        \centering
        \includegraphics[width=0.7\linewidth]{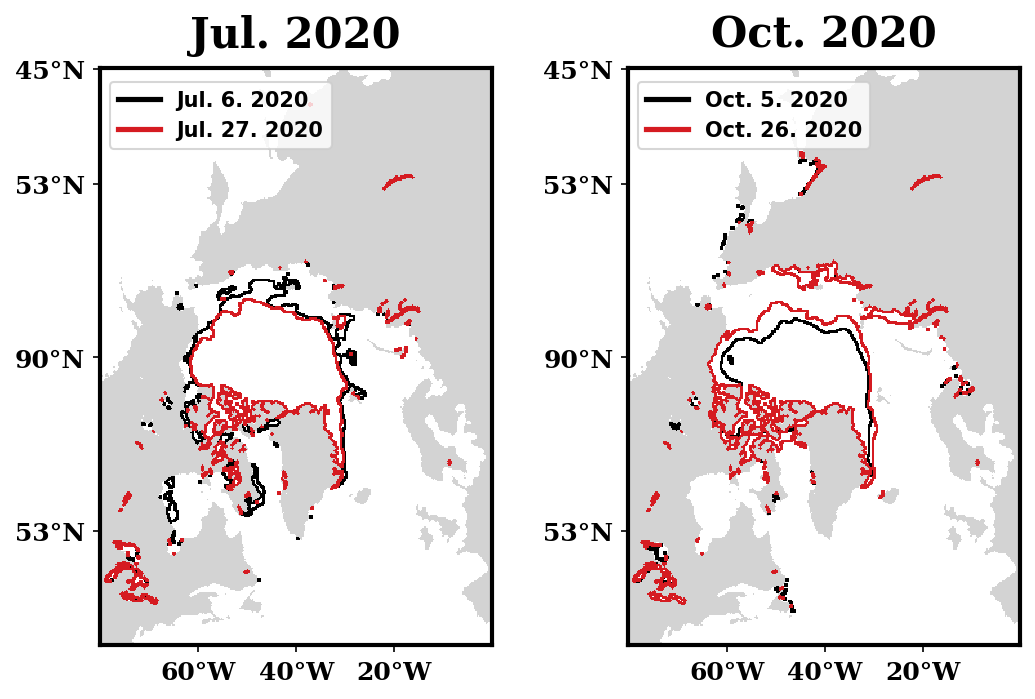}
        \caption{Sea ice edge shifts over time in July (left) and October 2020 (right).}
        \label{fig:sp_trend}
\end{figure}

Sea ice prediction involves two primary forecasts: sea ice concentration (SIC) at the pixel level and overall sea ice extent (SIE). In predicting sea ice accurately, it's crucial to capture the detailed dynamic changes in geographical morphology while retaining both spatial and temporal contexts. Temporally, through the time series of average SIC, we can effectively identify overall patterns of SIC. The average SIC time series from January to December, as shown in Figure \ref{fig:trend}, highlights a prominent trend, with its lowest point in August and peak sea ice concentration in December. Spatially, the patterns can be identified by visualizing the sea ice edge. As illustrated in Figure \ref{fig:sp_trend}, the sea ice edge in July undergoes inward during a decreasing trend or in October outward shifts during an increasing trend, while the interior remains relatively stable. It's worth noting that the sea ice forecaster models both temporal and spatial dynamics simultaneously.

Sea ice forecasting has relied on physical and statistical models designed to capture the inherent complexity of the phenomenon \cite{Wang2013SeasonalPO, collow2015improving, yuan2016arctic}. With recent advancements in deep learning, neural networks have emerged as a powerful tool capable of making predictions based on the inherent characteristics of the data \cite{Choi2019ArtificialNN, He2022AnIC, kim2023polargan}. IceNet \cite{Andersson2021SeasonalAS} is a notable example of this approach, successfully using deep learning to predict sea ice edges. Additionally, \cite{ren2022data} proposes integrating various spatial attention modules with deep learning models specifically for sea ice forecasting. 
However, these methods still have limitations in effectively addressing non-stationarity or trend shifts and in compartmentalizing the modeling of spatial and temporal dynamics like an architecture with spatial attention followed by temporal attention.

We suggest a framework of models aiming to overcome the limitations of existing sea ice prediction models in spatiotemporal modeling through three approaches. Firstly, we introduce neural ordinary differential equations (NODE) \cite{chen2018neural} and ConvNODE \cite{paoletti2019neural, li2021robust} to more efficient parameterization for the non-stationarity inherent in image time series. Our model extends the structure of ConvNODE to a neural network framework for spatiotemporal forecasting. Secondly, to mitigate information loss during the processing of time series components by the proposed model's encoder, we incorporate time series component decomposition into the data input layer. Prediction models utilizing the decomposition have enhanced performance by effectively extracting trends while retaining temporal information in predictions, particularly in non-stationary time series \cite{salles2019nonstationary}.
Finally, our model incorporates SIC images and ancillary data, similar to IceNet's use of climatic variables as supplementary data, further improving forecasting performance. 

Building upon our new framework, this study proposes a novel SIC forecasting model named Unicorn (U-Net for sea Ice forecasting using Convolutional OpeRation Node). By integrating the ConvNODE into the bottleneck of the U-Net structure, Unicorn utilizes the local spatial context within the images, all while preserving the temporal order of the data input. In addition to its distinctive model architecture, we enhance performance by incorporating ancillary image data pertaining to sea ice concentration. The main contributions of our work are as follows: 
\bed
    \item We propose an efficient deep learning architecture for fusing SIC image time series and ancillary static images to enhance forecasting accuracy. 
    \item We introduce a novel spatiotemporal latent modeling approach utilizing neural differential equations with convolutional operations.
    \item In real data analysis, our proposed model outperforms the state-of-art models, including domain-specific ones, in SIC and SIE forecasting from 1 to 4 weeks ahead.
\eed


\section{Related Work} \label{sec:rel}
This section presents various approaches to forecasting SIC, encompassing physical, statistical, and deep learning models. \cite{Wang2013SeasonalPO, collow2015improving} employed the Climate Forecast System, version 2 (CFSv2), which integrates atmospheric and oceanic dynamics alongside an interactive sea ice component, to forecast SIE. Nevertheless, these models relying on physics law entail significant computational expenses and often exhibit lower performance compared to statistical-based models. Statistical models, serving as an alternative to physical-based models, such as vector autoregression (VAR) \cite{Wang2016PredictingSA} and linear Markov model \cite{yuan2016arctic}, are utilized for forecasting SIC. However, statistical models have limitations regarding forecasting over mid or long terms due to their reliance on simplistic modeling approaches and the challenge of incorporating exogenous features.

Deep learning-based approaches for sea ice forecasting have been proposed to address these limitations. In initial studies, simple neural network architectures such as multilayer perceptron (MLP), recurrent neural networks (RNN), and CNN have been employed \cite{Chi2017PredictionOA, Song2018ARC}. However, these models encounter challenges in effectively capturing concurrent spatial and temporal features. To address the need for spatiotemporal representation learning and achieve higher accuracy, Convolutional LSTM (ConvLSTM) \cite{shi2015convolutional} and U-Net \cite{ronneberger2015u} architectures have been employed. 

ConvLSTM integrates Long Short-Term Memory (LSTM) and CNN layers to facilitate spatiotemporal learning \cite{huang2023interpretable}. In the context of daily SIC forecasting, ConvLSTM-based models have demonstrated superior performance compared to CNN-based models \cite{kim2019learning, Liu2021DailyPO}. Furthermore, \cite{Kim2021MultiTaskDL} proposes the multi-task ConvLSTM to forecast both SIC and SIE simultaneously. This multi-task model exhibits superior performance by leveraging the relationships between the two tasks. 
U-Net is initially designed as an encoder-decoder architecture for semantic segmentation for biomedical images. The encoder component maps inputs to lower-dimensional feature vectors by aggregating locally and globally spatiotemporal features. Subsequently, in the decoding step, these vectors are mapped back into the original feature space by exploiting the intermediate outputs of the encoder. Due to its superior performance in pixel-wise classification, U-Net has been applied to various domains beyond biomedical imaging, including weather forecasting \cite{seo2022simple, fernandez2021broad}, and traffic forecasting \cite{choi2020utilizing}. 
In SIC forecasting, \cite{Andersson2021SeasonalAS} proposes an ensemble architecture of U-Nets, IceNet, surpassing physics-based models' performance while maintaining computational efficiency. 
Additionally, \cite{ren2022data} proposed SICNet, a U-Net-based model designed for the weekly prediction of daily SIC, featuring various modules that enhance the capacity to capture spatiotemporal dependencies.

\section{Dataset and Preprocessing} \label{sec:data}
\begin{figure}[h]
    \centering
    \includegraphics[width=0.95\linewidth]{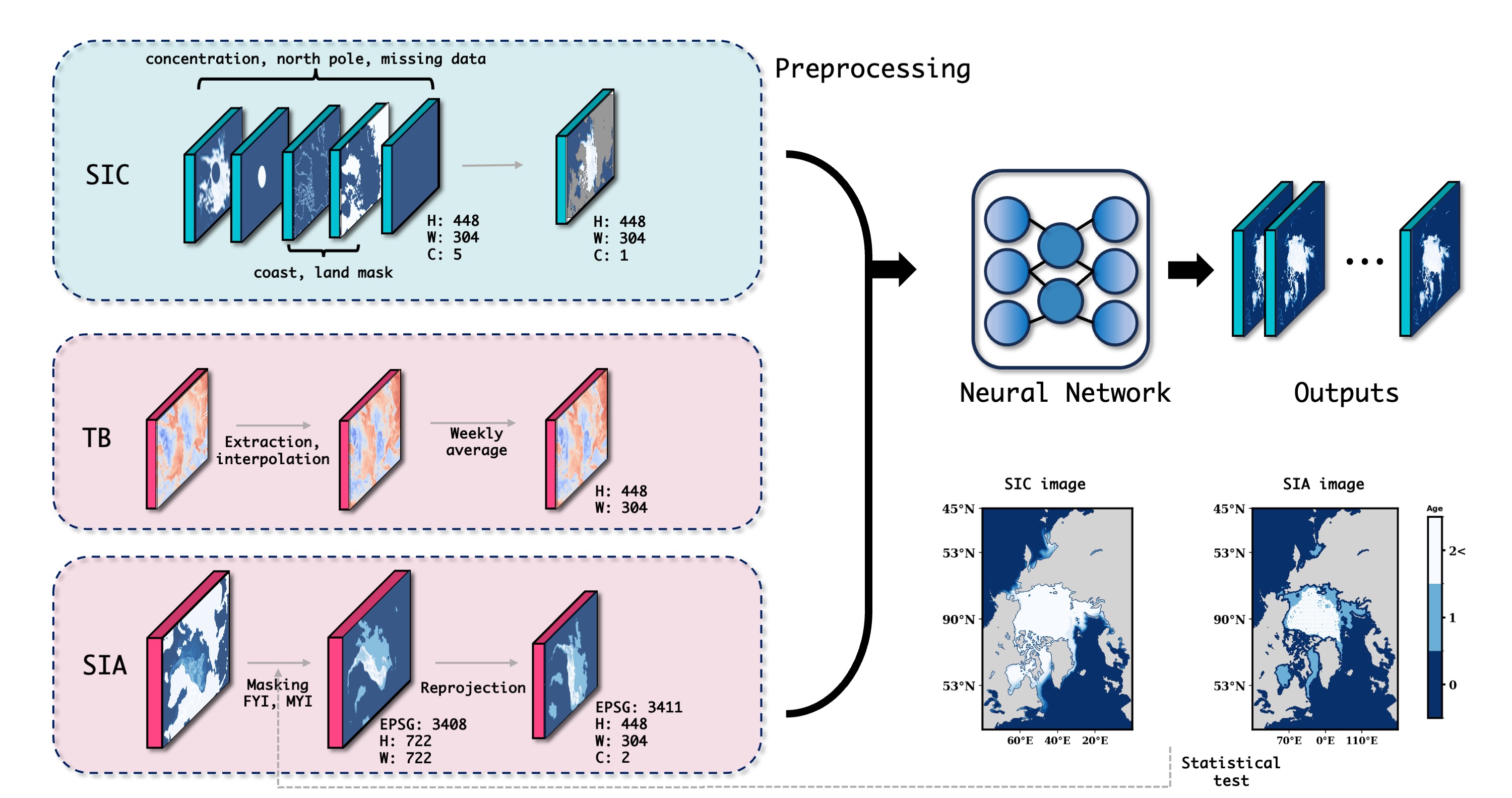}
    \caption{Preprocessing of our main {\color{teal} SIC} dataset and ancillary {\color{magenta} TB, SIA} datasets acquired from KOPRI and NSIDC. In the right-bottom, the SIC (left) and SIA (right) images are displayed to distinguish the distribution of sea ice based on its age. As shown in the SIA image, {\color{cyan}FYI} predominantly occupies the edges of the sea ice region.  
    The preprocessed data are then input into neural networks, where subsequent training and evaluation are conducted.}
    \label{fig:preprocess}
\end{figure}

This section demonstrates the datasets and preprocessing methods used in this study. We utilized three datasets: sea ice concentration (SIC), brightness temperature (TB), and sea-ice age (SIA) datasets. Among them, TB and SIA datasets are ancillary datasets that we selected to improve forecasting performance. 
The brightness temperature is one of the crucial factors in forecasting SIC because the SIC estimation algorithm is based on it \cite{Ivanova2015IntercomparisonAE}.
Furthermore, we utilized the SIA dataset to account for the significant differences in thickness and geographical location (spatial pattern) of sea ice depending on its age \cite{Chen2023CalibrationOU}. 
All datasets are characterized by image data, which have different sizes (height, width, and channels). Thus, TB and SIA images are resized to match SIC images. 
Figure \ref{fig:preprocess} visually presents all datasets and the entire data manipulation process preceding the forecasting model's training phase.
This preprocessing method in Figure \ref{fig:preprocess} ensures that both the spatial and temporal dimensions of the TB and SIA are optimally configured for integration with the SIC predictions.

\subsection{SIC dataset}
This study utilized observational sea-ice data from June 22, 1998, to June 
14, 2021. Weekly average SIC data mapped to a (25km $\times$ 25km) grid were provided by the Korea Polar Research Institute (KOPRI). The SIC images have a spatial resolution of 304$\times$448 and are gridded to a spatial reference system with an EPSG code of 3411. They consist of 5 image channels, including weekly SIC, North Pole, coastline, land mask, and missing data information. The weekly SIC, North Pole, and missing data information channels are combined into one channel, while the coastline and land mask channels are used to mask out continental regions.

\subsection{TB dataset}
The TB image dataset is sourced from the National Snow and Ice Data Center (NSIDC) \cite{Tschudi2019EASEGrid, Meier2022DMSP}. 
To align with our primary focus on forecasting SIC, we preprocessed the spatial and temporal resolution of TB dataset to match that of our SIC image dataset. The TB dataset is sourced from two different sensors on the Defense Meteorological Satellite Program (DMSP). The data from the Special Sensor Microwave/Imager (SSM/I) sensor on the F13 platform covers the period from 1995 to 2007, while the Special Sensor Microwave Image/Sounder (SSMIS) sensor on the F17 and F18 platforms covers from 2007 to the present. To achieve temporal coverage and gridded resolution (25km $\times$ 25km) consistent with the SIC data, we extracted TB at a 37GHz frequency channel that exists on both sensors. After extraction, we interpolated missing pixels using the nearest neighbor method, and the daily temporal resolutions of the TB dataset were averaged to match the weekly temporal resolution of the SIC dataset. 

\subsection{SIA dataset}
Figure \ref{fig:sp_trend} illustrates the decreasing patterns of SIC, particularly highlighting the high volatility at the edges. This underscores the importance of categorizing the edge and center areas in SIC forecasting. To effectively implement dynamic edge segmentation, we leverage the distinct characteristics of sea ice based on its age. 
First, we categorized sea ice into two binary groups: first-year (FYI) and multi-year sea ice (MYI). First-year sea ice, which is thicker than 30 cm but has not survived a summer melt season, is predominantly found along the edges of the sea ice region (as shown in the right-bottom of Figure \ref{fig:preprocess}). This type is highly volatile and prone to melting during the melt season. In contrast, multi-year sea ice, with thicknesses ranging from 2 to 4 meters and having survived at least one melt season, is mainly located in the central areas of the ice mass (as shown in the right-bottom of Figure \ref{fig:preprocess}) and exhibits minimal seasonal change. 
Due to these pronounced characteristics, which significantly impact sea ice dynamics, we selected to incorporate the SIA dataset as ancillary data in our analysis. Additionally, through statistical testing such as ANOVA, we validated the categorization into two groups. The results indicate a significant difference in SIC between the two groups, with a $p$-value $< 0.01$, confirming the validity of our approach to provide more information on sea ice.

Thus, for our SIC forecasting task, we categorized the pixel values (1 to 16) from the SIA images into two groups: age $= 1$ and age $\geq 2$.
The SIA dataset with a spatial resolution of $722 \times 722$ and mapped to a (12.5km $\times$ 12.5km) grid under the spatial reference system EPSG code 3408, requires a coordinate system transformation to convert its pixel coordinates into geospatial coordinates. Subsequently, we performed a reprojection to align the geospatial coordinates of the SIA dataset with those of the SIC dataset, standardizing the spatial resolution to $304 \times 408$ mapped to a (25km $\times$ 25km) grid under the spatial reference system EPSG code 3411. This ensures consistency in spatial analysis across datasets.

\begin{figure}[!t]
    \centering
    \includegraphics[width=0.9\linewidth]{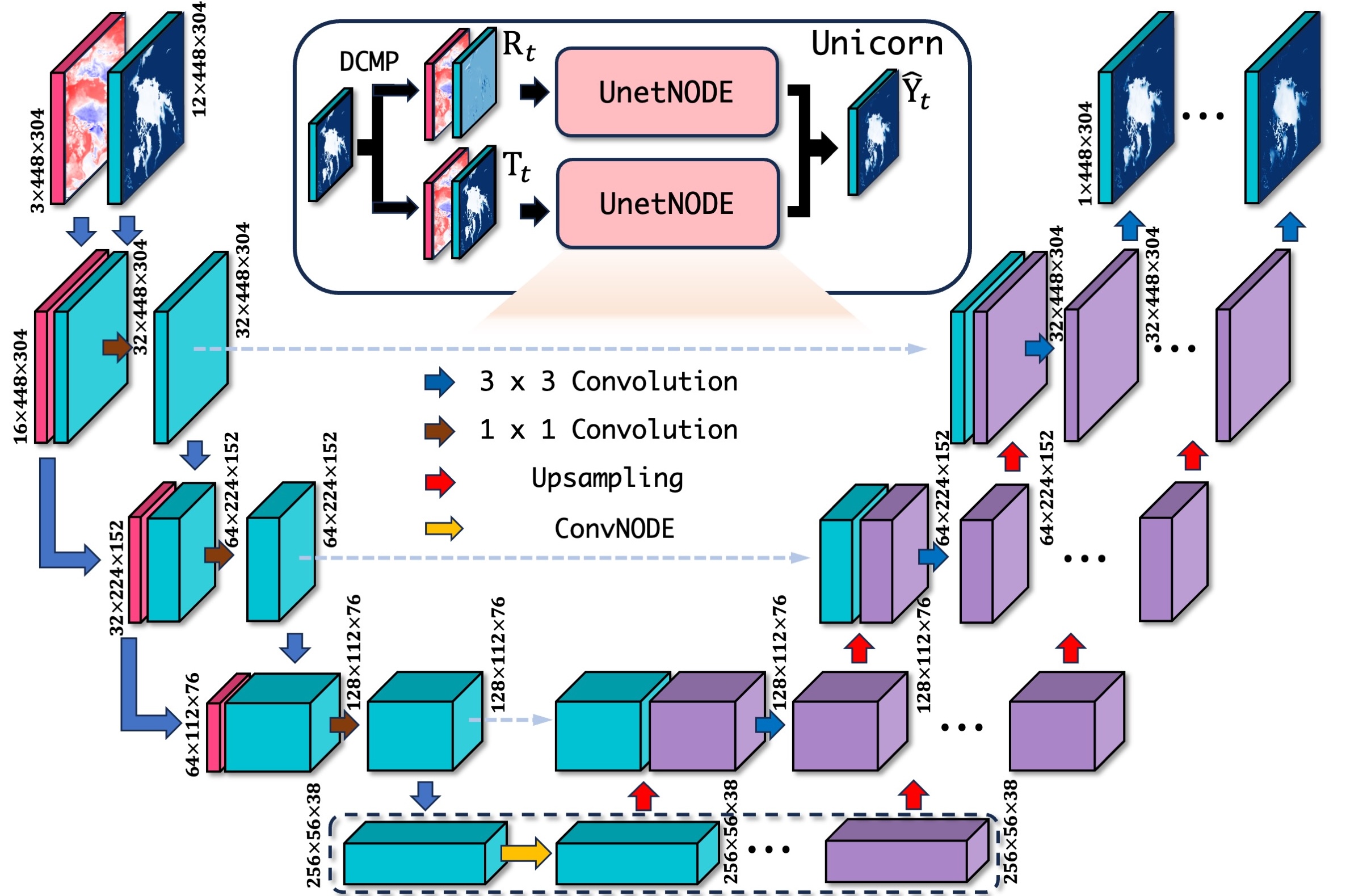}
    \caption{Overall architecture of our proposed model, Unicorn, in the middle above, and UnetNODE (example for trend component $\mathbf{T}_t$). {\color{teal} Teal} and {\color{magenta} red}-colored boxes represent intermediate outputs of the encoder, while {\color{violet}violet}-colored boxes represent intermediate outputs of ConvNODE and decoder.}
    \label{fig:arch}
\end{figure}

\section{Proposed method} \label{sec:prop}

This section introduces the proposed method, Unicorn, which employs the U-Net architecture and convolutional operations as the rate of change function of latent variables. The network architecture of Unicorn is displayed in Figure \ref{fig:arch}. 
Unicorn consists of two UnetNODEs, each comprising three components: the encoder, ConvNODE, and decoder. Each component is detailed in the subsequent sections. Our proposed model aims to forecast the sea ice concentration as a time series of length $\tau$, measured in weeks. The model takes SIC image from time $(t-L+1)$ to $t$, denoted as $\bX_t \in \mathbb{R}^{L \times (H \times W)}$. Note that $L$ represents the sequence length with consideration of the channel dimension. The dimensions $H$ and $W$ refer to the height and width of each image, respectively. Additionally, UnetNODE takes ancillary datasets as inputs, the TB 
and SIA, as denoted $\mathbf{B}_t \in \mathbb{R}^{1 \times (H \times W)}$ and $\mathbf{S}_t  \in \mathbb{R}^{2 \times (H \times W)}$, respectively. Because we consider each group of SIA described in Section \ref{sec:data} as a binary variable, $\mathbf{S}_t$ has two channels (age $= 1$ and age $\geq 2$).  
The model output consists of sea ice concentration predictions from $(t+1)$ to $(t+\tau)$ time points, denoted as $\hat{\bY}_t \in [0,1]^{\tau \times (H\times W)}$, while the corresponding target SIC is denoted as $\bY_t \in [0,1]^{\tau \times (H\times W)}$. For our experiment, we set $L$ and $\tau$ to $12$ and $4$, respectively. In other words, our model forecasts the sea ice concentration for $4$ weeks ahead using information from the preceding $12$ weeks.

\subsection{Decomposition}
Firstly, based on the results of an explanatory data analysis in Figure \ref{fig:trend}, we decompose the input image into two components: trend and residual. Time series decomposition has enhanced forecasters efficiently \cite{salles2019nonstationary}. In our image data, the trend component of $\bX_t$, denoted as $\bT_t$, is easily extracted through the utilization of a moving average filter ($MA_K$) with a size of $K$, an odd number less than $L$, as follows:
\begin{align*}
\mathbf{T}_t = MA_K(tpad(\bX_t)), 
\end{align*}
where $\mathbf{T}_t(l,i,j) = \frac{1}{K} \sum_{k=- \lfloor K/2 \rfloor }^{\lfloor K/2 \rfloor} tpad(\bX_t)(l+k,i,j)$, and $tpad(\cdot)$ represents a padding technique to preserve the time length, while $\mathbf{M}(l, i, j)$ denotes the $(i,j)$ pixel value of the $l$-th image in $\mathbf{M}$. \\ 
The residual component, $\mathbf{R}_t$, is calculated by detrending the original input $\bX_t$ as follows: $\mathbf{R}_t = \bX_t - \bT_t$. Then, the components, $\bT_t$ and $\mathbf{R}_t$, are each utilized as input in their own U-Net architecture, as described in Figure \ref{fig:arch}. In other words, we employ two separate U-Nets for each component to forecast in a parallel fashion. 
This step of separating the image data into trend and residual components is referred to as DCMP.
From now on, since the process for the residual component is identical, we will only explain the process for the trend component.

\subsection{Encoder}
Our proposed method is designed to integrate multiple pieces of information from various datasets. 
At the time point $t$, the encoder takes three images: the component of the SIC ($\mathbf{T}_t$ or $\mathbf{R}_t$), the TB ($\mathbf{B}_t$), and the SIA ($\mathbf{S}_t$). 
Because the channels of SIC and ancillary images have different semantics (time points in SIC and dimensions of variables in ancillary images), the contracting path of the encoder is split into two paths: the main path and the ancillary path. 
The main path takes the components of the SIC, and its architecture follows the typical type of the U-Net encoder. 
 The main path comprises recursively stacked blocks combined by $3 \times 3$ convolutions, batch normalization, ReLU activation function, and $2\times2$ max pooling. 
The ancillary path takes the concatenated ancillary images, $[\mathbf{B}_t; \mathbf{S}_t] \in \mathbb{R}^{3\times(H \times W)}$ and consists of the same blocks as the main path. 
The output of the main path is integrated with that from the ancillary path using a $1 \times 1$ convolution.  
Consequently, the encoder seamlessly fuses multiple images, and the recursive fusions via $1 \times 1$ convolutions provide additional information to the upsampling process by concatenating copied features. 
Let the output of the encoder be $\bz_{t}(0)$ at time point $t$, then the process in the encoder can be written as follows:
\begin{align*}
&\mathbf{A}_t^{(1)} = Convblock^{(1)}(cat(\mathbf{B}_t, \mathbf{S}_t)) \\ 
&\mathbf{T}_t^{(1)} = Convblock^{(2)}(\mathbf{T}_t) \\ 
&\mathbf{F}_t^{(1)} = Conv^{(1)}_{1 \times 1}(cat(\mathbf{A}_t^{(1)}, \mathbf{T}_t^{(1)}))\\ 
&\mathbf{A}_t^{(2)} = Convblock^{(3)}(\mathbf{A}_t^{(1)}) \\
&\mathbf{F}_t^{(2)} = Conv^{(2)}_{1 \times 1}(cat(Convblock^{(4)}(\mathbf{F}_t^{(1)}), \mathbf{A}_t^{(2)})) \\ 
&\mathbf{A}_t^{(3)} = Convblock^{(5)}(\mathbf{A}_t^{(2)}) \\
&\mathbf{F}_t^{(3)} = Conv^{(3)}_{1 \times 1}(cat(Convblock^{(6)}(\mathbf{F}_t^{(2)}), \mathbf{A}_t^{(3)})) \\
&\bz_{t}(0) = Convblock^{(7)}(\mathbf{F}_t^{(3)}),    
\end{align*}
where $Conv_{i\times i}$ is the convolutional layer with an $i \times i$ kernel, and $Convblock$ is a block composed of two [$Conv_{3\times 3}$, batch normalization, relu] operations. Additionally, $cat$ is a concatenation operator along the time dimension.

\subsection{ConvNODE}
ConvNODE effectively captures the intricate spatiotemporal dynamics in sea ice images extracted through the encoder. 
It predicts the temporal evolution of the latent variable $\bz_t(\kappa)$ with $\kappa > 0$, derived from each component $\bT_t$ and $\mathbf{R}_t$, at the output image's timestamps using NODE framework, where the rate of change function is represented by the convolutional operation. 

The theoretical foundation enabling this structure is as follows. 
By representing ODE as differential equations dependent on time $k$ and solving them, we can derive changes in the variable over time. Let $\kappa$ be the current time point and $\kappa+1$ be the next time point. The rate of change function for $\bz_t(\kappa)$ is denoted by $f(\bz_t(k))$. Based on these definitions, the temporal evolution of $\bz_t(\kappa)$ can be calculated using ODE as follows:
\bea \label{eq: ode}
\bz_t(\kappa+1) = \bz_t(\kappa) + \int_{\kappa}^{\kappa+1} f(\bz_t(s)) ds.
\eea

By defining \eqref{eq: ode}, we establish a strong foundation for predicting the intricate dynamics of latent variables. The effectiveness of applying the NODE framework relies on selecting the appropriate rate of change function, $f$, for a given purpose. Based on this understanding, \cite{lim2023long} defines the function $f$ as an affine function, such as $f(\bz_t(s)) = \bW\bz_t(s) + \mathbf{b}$, for time series forecasting of tabular datasets. 
To proficiently model the temporal dynamics of image sequences, we employ convolutional operations instead of a linear $f$. It is recognized for its ability to adeptly encapsulate both spatial and temporal information within image and video datasets \cite{simonyan2014two}.
Aligned with the ODE approach, the temporal dynamics of $\bz_t(\kappa)$ as a function over time are defined as follows:
\bea \label{eq: conv_node}
\bz_t(\kappa+1) = \bz_t(\kappa) + \int_{\kappa}^{\kappa+1} Conv_{3 \times 3}(\bz_t(k)) dk.
\eea 

Through ConvNODE in \eqref{eq: conv_node}, the latent variables are computed and passed to the decoder as inputs. This structure effectively combines the advantages of NODE and convolutional operations. Therefore, it accurately reflects spatiotemporal variations in latent variables extracted from sea ice and ancillary images. 

\subsection{Decoder}
The decoder utilizes the outputs, $\bz_{t}(0), \dots, \bz_{t}(\tau-1)$, to forecast the future SIC from $(t+1)$ to $(t+\tau)$ time points, denoted as $\bY_{t}$. 
Contrary to the encoder, which has only two paths (main and ancillary paths), the decoder consists of $\tau$ expansive paths to predict $\tau$ SIC images by leveraging the corresponding latent variable $\bz_{t}(\kappa), \kappa=0, \dots, \tau-1$. 
Contextual information within $\bz_{t}(\kappa)$ is expanded by $2 \times 2$ upsampling convolutions in each expansive path. 
Subsequently, it is combined with the output of the encoder through the enhanced skip connection (copy and concatenating) with ancillary information. 
After three upsampling processes, the output image has the same size as the original input, as follows : 
\begin{align*}
&\mbox{For } \kappa = 0, \dots, \tau-1, \\ 
& \bz_{t}^{(1)}(\kappa) = Convblock^{(8)}_{\kappa}(cat(upsample^{(1)}_{\kappa}(\bz_{t}(\kappa)), \mathbf{F}_t^{(3)})) \\
& \textbf{z}_{t}^{(2)}(\kappa) = Convblock^{(9)}_{\kappa}(cat(upsample^{(2)}_{\kappa}(\bz_{t}^{(1)}(\kappa)), \mathbf{F}_t^{(2)})) \\
& \bz_{t}^{(3)}(\kappa) = Convblock^{(10)}_{\kappa}(cat(upsample^{(3)}_{\kappa}(\bz_{t}^{(2)}(\kappa)), \mathbf{F}_t^{(1)})) \\
& \mathbf{\Tilde{T}}_t = Conv^{(1)}_{3 \times 3}(cat(\bz_{t}^{(3)}(0), \cdots, \bz_{t}^{(3)}(\tau-1))),
\end{align*}
where $upsample$ is an operation to increase the spatial resolution of the input, performed using transposed convolution. $\sigma(\cdot)$ denotes the sigmoid activation function for scaling outputs. \\ 
Simultaneously applying the entire process from the encoder to the decoder with $\mathbf{R}_t$, we obtain $\Tilde{\mathbf{R}}_t$, and subsequently, the predictions are calculated as $\hat{\mathbf{Y}}_t = \sigma(\mathbf{\Tilde{T}}_t + \mathbf{\Tilde{R}}_t)$.

\subsection{Training}
This section introduces a training process for our proposed model. Many existing approaches, such as \cite{Ali2022MTIceNetA, ren2023predicting}, utilized the MSE as a loss function for the prediction tasks.  
Because the SIC takes values in $[0, 1]$ like a normalized pixel intensity, the binary cross-entropy (BCE) can be used for a dissimilarity criterion between two images \cite{creswell2017denoising}. 
Moreover, \cite{de2022novel} demonstrated that models minimizing the BCE tend to outperform others that minimize different loss functions, such as a mean squared error (MSE), particularly in tasks related to image reconstruction. Given the target $\bY_t$ and output $\hat{\bY}_t$, our model is instructed to minimize the BCE loss $\mathcal{L}$ as follows: 
\begin{align*}
\mathcal{L}(\Theta; \mathcal{T}) =
\sum_{t \in \mathcal{T}} \sum_{l=1}^\tau\sum_{i,j} \bY_{t}(l, i,j)\log\hat{\bY}_t(l,i,j;\Theta) + \\  (1-\bY_{t}(l, i,j))\log(1-\hat{\bY}_t(l,i,j;\Theta)), 
\end{align*}
where $\Theta$ is the entire parameters of the model, $\mathcal{T}$ is a time point set of the training dataset, and $\bY_t(l, i,j)$ and $\hat{\bY}_t(l, i,j)$ denote $(i,j)$ element of target and predicted image for $(t+l)$ time point, respectively. The model parameters $\Theta$ can be updated by stochastic gradient descent-based algorithms. 

\section{Experiment} \label{sec:ex} 
\begin{figure}[!t]
    \centering
    \includegraphics[width=0.9\linewidth]{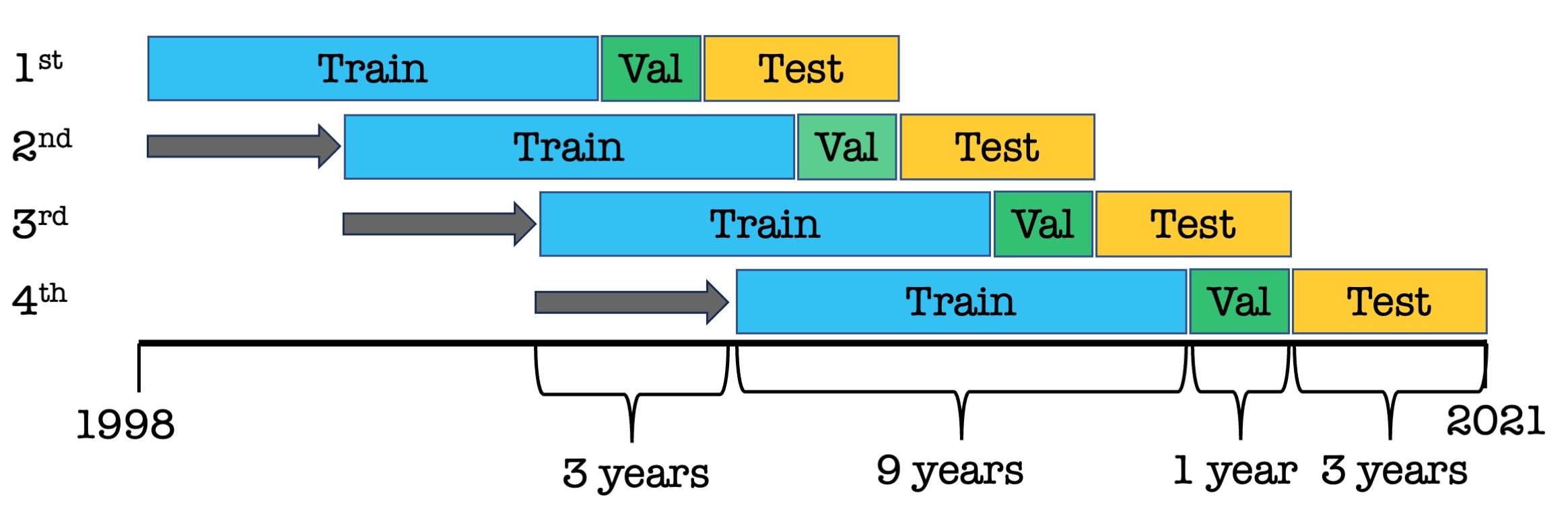}
    \caption{Sliding window evaluation procedure. The total dataset spans from 1998 to 2021 and is divided into four overlapping periods, each fifteen years (training: 9 years, validation: 1 year, test: 3 years) in length (denoted as $1\textsuperscript{st} \sim 4\textsuperscript{th}$).}
    \label{fig:swp}
\end{figure}

This section introduces our evaluation procedure and metrics used in our experiment and presents performance comparisons between the proposed model and existing ones. To ensure reliable results, we implemented a time series cross-validation strategy using our dataset of SIC, TB, and SIA images from 1998 to 2021. The data are divided into four overlapping fifteen-year periods, and each new period begins three years after the previous one. In each segment, the first 11 years are designated for training, the 12th year for validation, and the last three years for testing. This sliding window approach allows for comprehensive coverage and utilization of the dataset over time. Figure \ref{fig:swp} presents our sliding window procedure for the model evaluation in this study.  

For training and validation, batch sizes are set to 8, while a batch size of 1 is used for the test set to facilitate detailed performance analysis. The \texttt{AdamW} optimizer \cite{Loshchilov2017DecoupledWD} with a learning rate of 0.001 is used, and an early stopping mechanism is applied if no improvement in validation loss is observed for 50 iterations. All implementations, including benchmarks, were conducted using \texttt{PyTorch} on an A10 GPU and CentOS Linux. The source code is publicly accessible at \url{https://github.com/Optim-Lab/sif-models}. 

\subsection{Metrics} \label{sec:metric}
This study assesses sea ice forecasting models through two tasks: SIC and SIE forecasting. 
For SIC forecasting, mean absolute error (MAE) and root mean squared error (RMSE) are selected as evaluation metrics. These metrics are widely used for assessing the accuracy of predicted sea ice concentrations at individual pixels in the forecasted images. All metrics are computed using pixel values within non-land ice areas to ensure a precise evaluation specifically targeting sea ice. We denote $\mathcal{N}$ as the set of pixel coordinates representing non-land ice areas in the images. Thus, MAE and RMSE between the ground truth ($\bY$) and predicted value ($\hat{\bY}$) are computed as follows: 
\begin{align*}
& \mbox{MAE} = \frac{1}{\tau N|\mathcal{T}'|} \sum_{t\in\mathcal{T}'} \sum_{l=1}^\tau\sum_{(i,j) \in \mathcal{N}}\left\| \mathbf{Y}_t(l,i,j) - \hat{\mathbf{Y}}_t(l,i,j) \right\|_1 \\ 
& \mbox{RMSE} = \sqrt{\frac{1}{\tau N|\mathcal{T}'|} \sum_{t\in\mathcal{T}'}\sum_{l=1}^\tau\sum_{(i,j) \in \mathcal{N}} \left\| \mathbf{Y}_t(l,i,j) - \hat{\mathbf{Y}}_t(l,i,j) \right\|_2^2},    
\end{align*}
where $\mathcal{T}'$ represents the time point set of the test dataset, and $N$ denotes the cardinality of $\mathcal{N}$, i.e., $|\mathcal{N}|$.

For SIE forecasting, considering the critical importance of accurately predicting sea ice areas, we utilize the integrated ice-edge error (IIEE) metric. We also employ the F1-score, accuracy, and mean intersection over union (mIoU), which are widely used in computer vision tasks. To evaluate SIE forecasting performance, both ground truth and predicted SIC images are transformed into binary images using a threshold of $0.15$. Let $\mathsf{Y}_t$ and $\hat{\mathsf{Y}}_t$ represent the binary images transformed from $\bY_t$ and $\hat{\bY}_t$, respectively, where $\mathsf{Y}_t, \hat{\mathsf{Y}}_t \in \{0, 1\}^{\tau \times (H \times W)}$. The metrics for SIE forecasting are computed as follows:
\begin{align*}
& \mbox{IIEE} = \frac{1}{\tau|\mathcal{T}'|}\sum_{t\in\mathcal{T}'}\sum_{l=1}^\tau\sum_{(i,j) \in\mathcal{N}} \left\| \mathsf{Y}_t(l,i,j) - \hat{\mathsf{Y}}_t(l, i,j) \right\|_1 \\
& \mbox{IoU}_t = \frac{1}{\tau}\sum_{l=1}^\tau \frac{|\{(i,j)| \mathsf{Y}_t(l,i,j) = 1\} \cap \{(i,j)| \hat{\mathsf{Y}}_t(l,i,j) = 1\}|}{|\{(i,j)| \mathsf{Y}_t(l,i,j) = 1\} \cup \{(i,j)| \hat{\mathsf{Y}}_t(l,i,j) = 1\}|} \\ 
& \mbox{mIoU} = \frac{1}{|\mathcal{T}'|}\sum_{t \in \mathcal{T}'} \mbox{IoU}_t, ~~~~~ \mbox{F1} =  \frac{2 \cdot \text{TP}}{2 \cdot \text{TP} + \text{FP} + \text{FN}}, 
\end{align*}
where TP, FP, and FN denote the number of true positives, false positives, and false negatives, respectively.

\subsection{Compared Models}
For comparative analysis, we evaluate seven benchmark models, including domain-specific ones, alongside our proposed model. Each baseline model retains the structure and hyperparameter configurations of its original version, with adjustments to facilitate the generation of $\tau$ prediction images from $L$ input images. Additionally, we applied zero padding to convolutional operations wherever necessary to maintain the size of the images. The list of benchmark models and their summaries are as follows:     
\subsubsection{Baseline models} 
CNN has been an effective baseline model for various image-related tasks and is often used in ice concentration prediction studies as well. In this experiment, CNN is based on a model structure from previous research dedicated to predicting ice concentrations \cite{Kim2021MultiTaskDL}.

ConvLSTM combines effective convolution operations for feature extraction from image data with the powerful LSTM for time-series prediction. This model is primarily utilized for predicting changing image sequences and is frequently employed as a baseline in ice concentration prediction research. 

U-Net \cite{ronneberger2015u} has established itself as an excellent baseline model with its innovative structure, achieving significant success in the field of medical image segmentation. Subsequently, it has been applied across various domains, including classification, enhancement, and forecasting, demonstrating notable achievements. Particularly, as mentioned in Section \ref{sec:rel}, many ice concentration prediction models have utilized U-Net as their baseline model. 

DU-Net and NU-Net are specifically designed for comparison experiments in this study. While U-Net serves as a robust baseline with its encoder-decoder structure for sea ice prediction, concerns are raised regarding its capacity to effectively capture temporal information. To address this, we augmented U-Net with a time-series decomposition structure. We incorporated two simple time-series decomposition methods proposed in \cite{Zeng2022AreTE}. The trend and residual components decomposed by the decomposition block are then processed through the U-Net structure, combined, and passed through the sigmoid function to form the final prediction results like our proposed model.

\subsubsection{Domain-specific models}
SICNet \cite{ren2022data} enhanced the U-Net architecture by integrating spatiotemporal attention mechanisms into its convolution layers. This enhancement includes SICNet (CBAM), which combines channel and spatial attention, and SICNet (TSAM), where temporal attention replaces channel attention. These models demonstrated superior forecasting performance compared to traditional CNN and LSTM approaches in forecasting SIC images over a seven-week period based on seven weeks of SIC images. In this experiment, the hyperparameters are adjusted to align with our problem.

\begin{table}[b]
\caption{Experimental results of forecasting models. The most favorable value is highlighted in bold.}\label{tab:results} 
\centering
\begin{tabular}{lccccc}
\toprule
 Task & \multicolumn{2}{c}{SIC} & \multicolumn{3}{c}{SIE}\\
 \cmidrule(lr){2-3} \cmidrule(lr){4-6}
 Metric & MAE $\downarrow$ & RMSE $\downarrow$ & IIEE $\downarrow$ & mIoU $\uparrow$& F1 $\uparrow$ \\ 
\midrule
CNN & 0.031 & 0.096 & 1975.711 & 0.879 & 0.935\\ 
ConvLSTM & 0.035 & 0.096 & 2124.798 & 0.873 & 0.932 \\ 
U-Net & 0.024 & 0.077 & 1428.064 & 0.911 & 0.953 \\ 
DU-Net & 0.025 & 0.075 & 1409.271 & 0.912 & 0.953 \\  
NU-Net & 0.024 & 0.076 & 1401.810 & 0.912 &  0.954\\ 
SICNet (CBAM)  & 0.024 & 0.074 & 1374.186 & 0.915 & 0.955 \\ 
SICNet (TSAM) & 0.024 & 0.075 & 1402.616 & 0.913 & 0.954 \\ 
Unicorn & \textbf{0.023} & \textbf{0.071} & \textbf{1270.477} & \textbf{0.920} & \textbf{0.958} \\ 
\botrule
\end{tabular}
\end{table}

\subsection{SIC forecasting performance}

For the SIC forecasting task, we assess both benchmarks and Unicorn using two metrics, MAE and RMSE, as defined in Section \ref{sec:metric}. Performance results on the test dataset are presented in Table \ref{tab:results}, demonstrating that Unicorn outperforms other models in both metrics. Specifically, Unicorn exhibits significant superiority over all benchmarks in the MAE metric, which is known for its robustness against outliers, achieving an average improvement of 12.11\% (with an improvement of 4.17\% in comparison to the next best model). In evaluation using the RMSE, Unicorn achieved an average 11.60\% improvement versus other models with an improvement of 4.05\% compared to SICNet (CBAM).


\begin{figure}[t]
    \centering
    \includegraphics[width=0.7\linewidth]{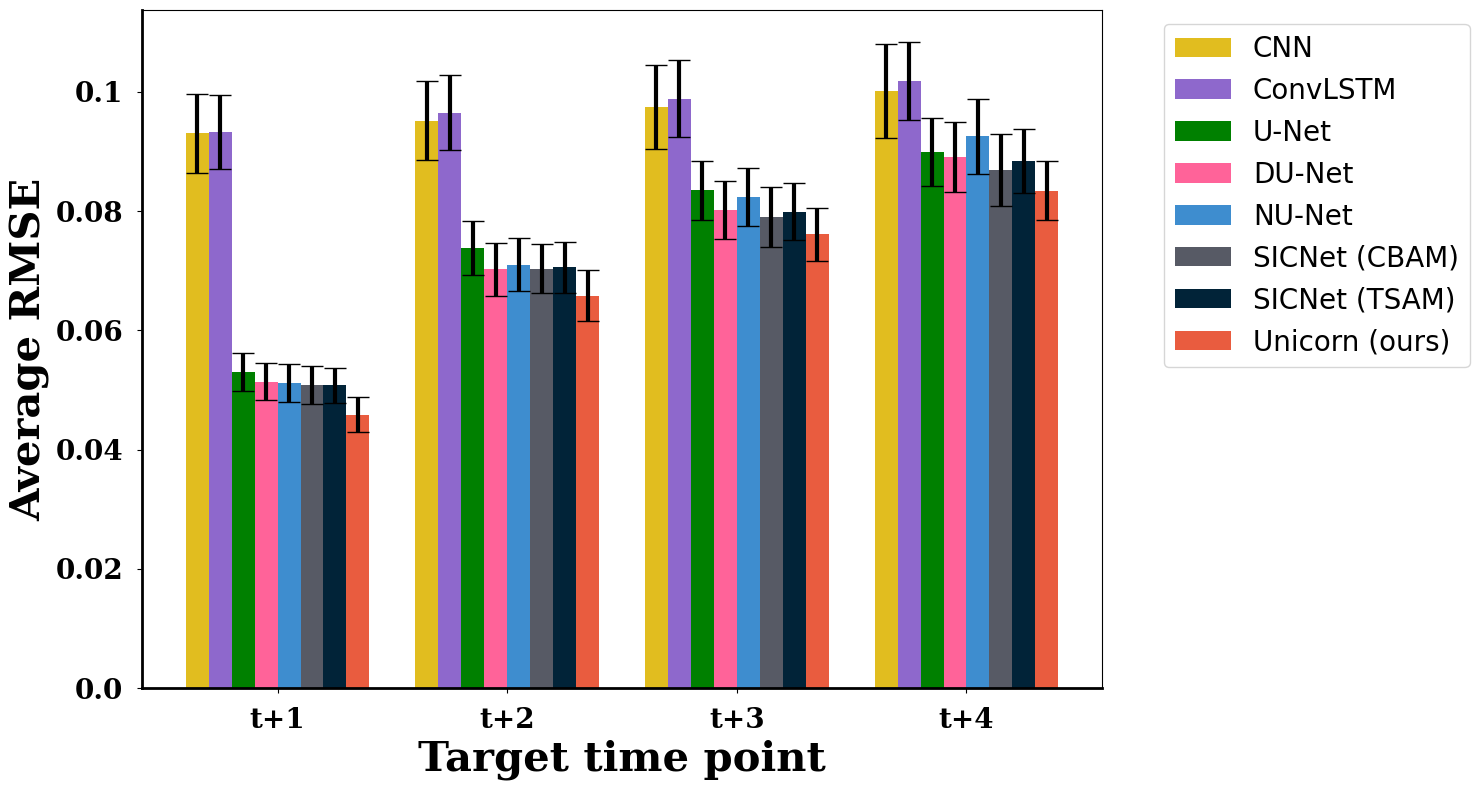}
    \includegraphics[width=0.7\linewidth]{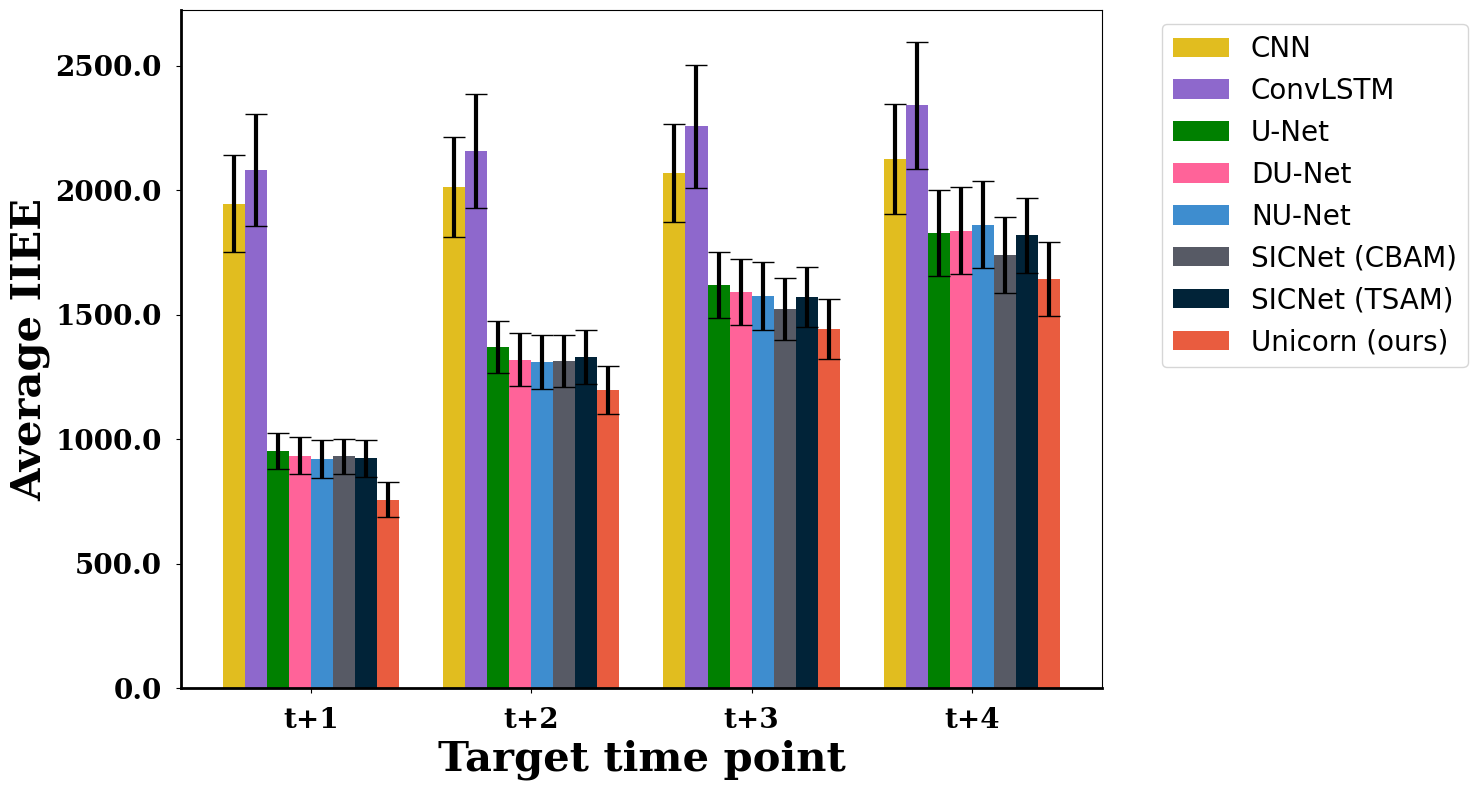}
    \caption{Average RMSE (top) and IIEE (bottom) with its interval, average $\pm 1$ standard error, of forecasting models across different target time points.}
    \label{fig:bar_charts}
    
\end{figure}

\begin{figure}[t]
    \centering
    \includegraphics[width=0.7\linewidth]{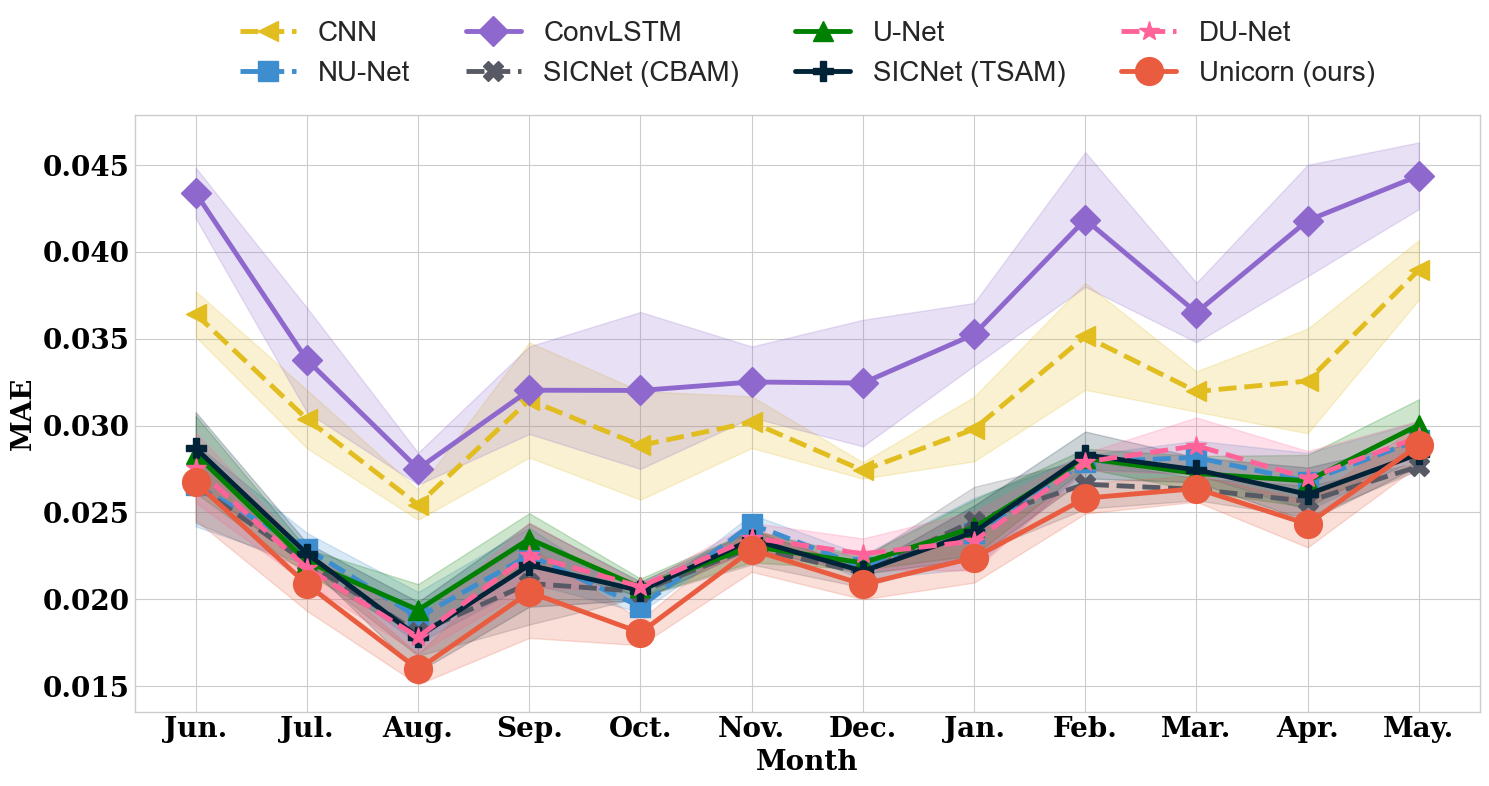}
    \includegraphics[width=0.7\linewidth]{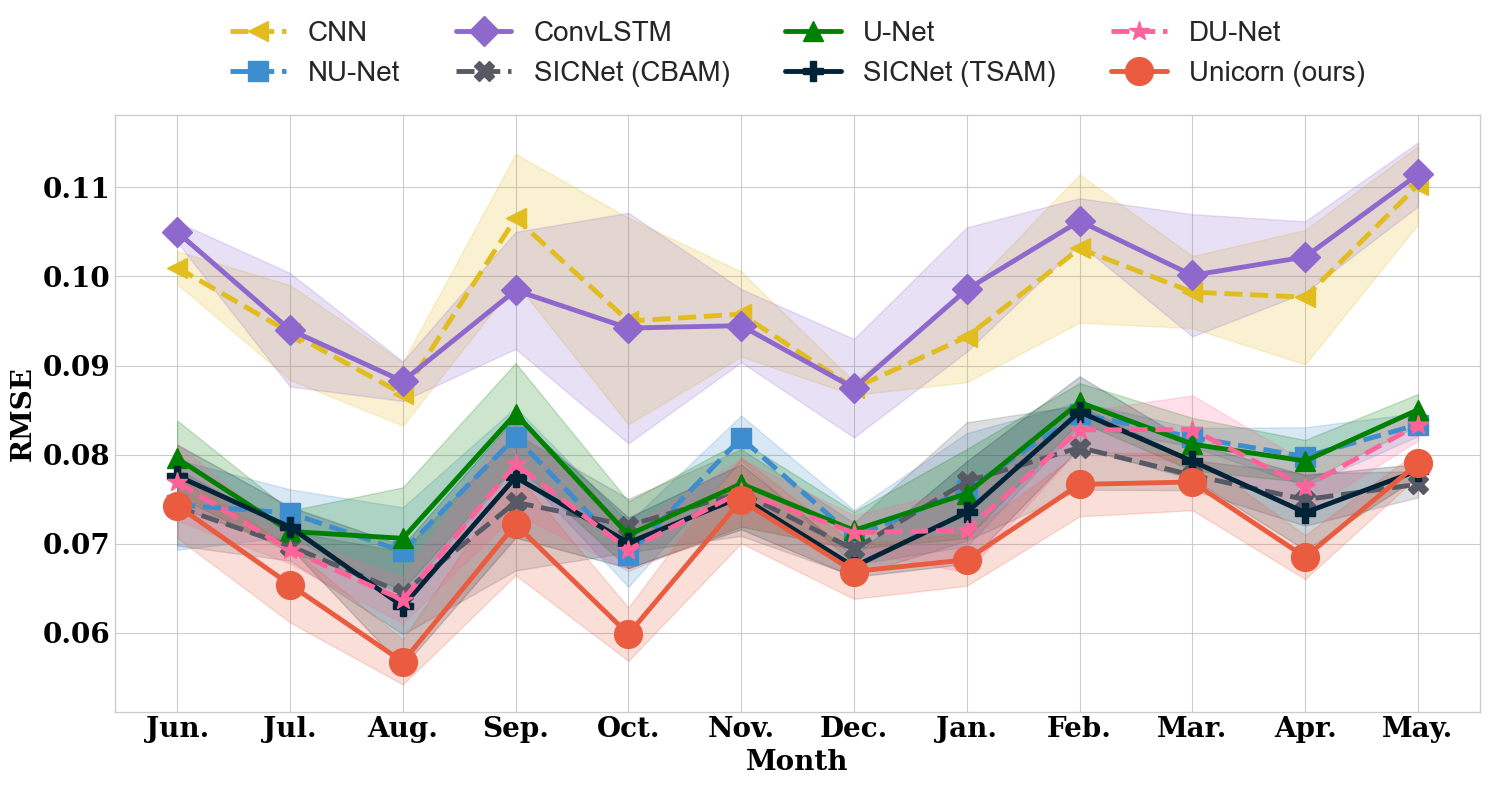}
    \caption{Monthly average SIC forecasting performance of Unicorn and compared models from June 2020 to May 2021.}
    \label{fig:sic_monthly_result}
\end{figure}


\begin{figure}[!t]
    \centering
    \subfigure{\includegraphics[width=0.9\linewidth]{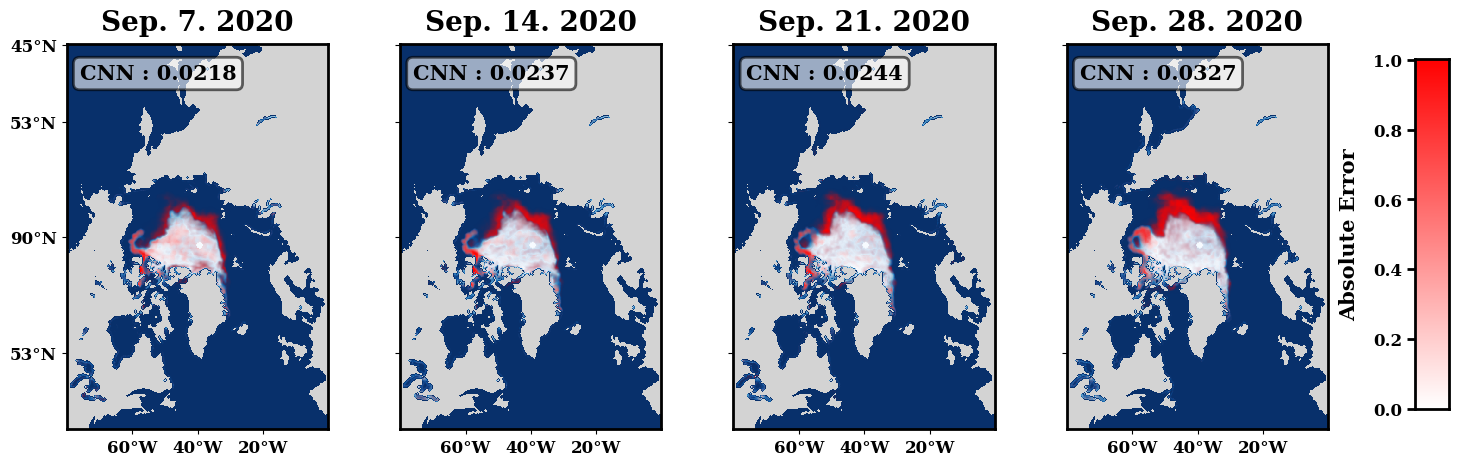}}
    \subfigure{\includegraphics[width=0.9\linewidth]{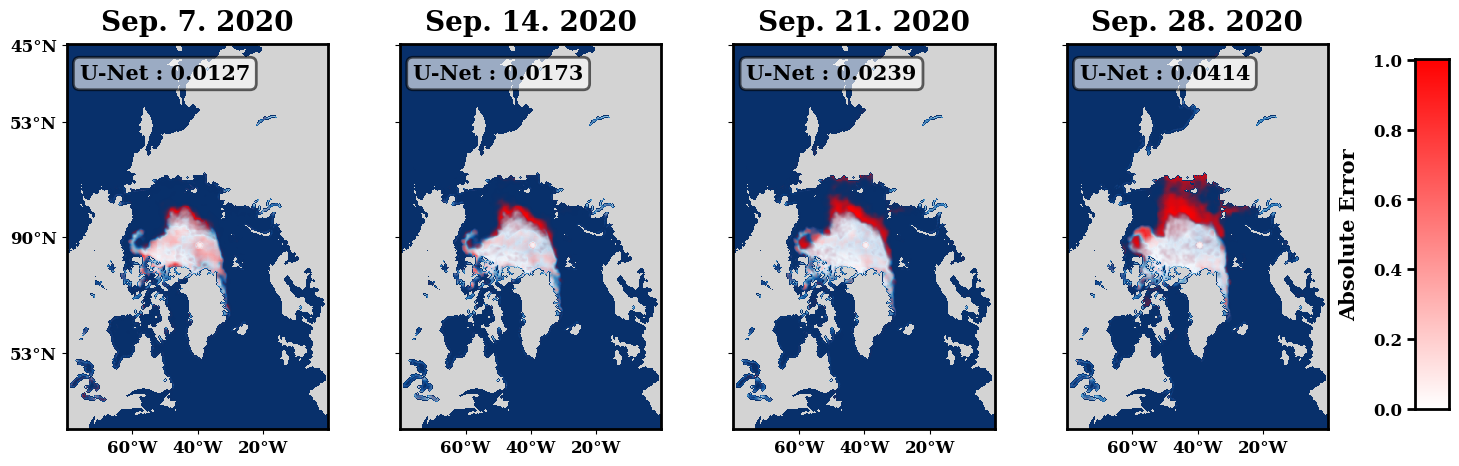}}
    \subfigure{\includegraphics[width=0.9\linewidth]{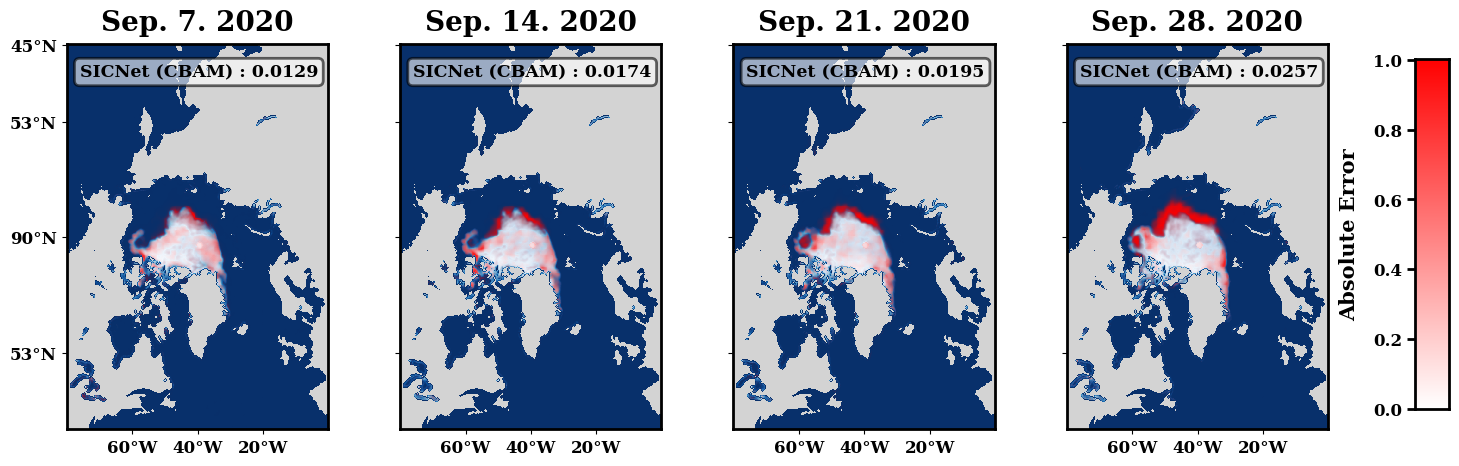}}
    \subfigure{\includegraphics[width=0.9\linewidth]{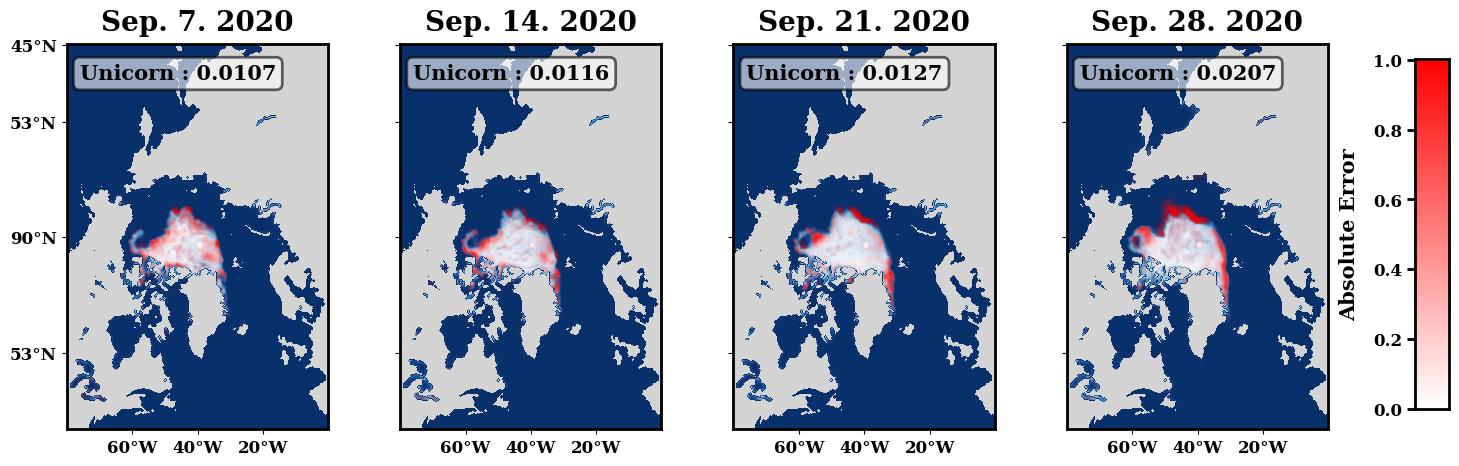}}
    \caption{
    SIC forecasting results. From the top row to the bottom row, the images showcase the results from CNN, U-Net, SICNet (CBAM), and Unicorn, along with their MAE values. The \textcolor{red}{red} color represents the absolute error between prediction and ground truth.
    }
    \label{fig:sic_result}
\end{figure}

Figure \ref{fig:bar_charts} (top) displays the average RMSE across different target time points. There is a significant difference between the U-Net-based models and the others. CNN and ConvLSTM consistently show degraded performance irrespective of the target time points.
In contrast, the U-Net-based models (from U-Net to Unicorn) exhibit competitive performance in one-step ahead forecasting but declining performance as the target time points extend.
Among them, Unicorn has the lowest average errors at each target time point.

Figure \ref{fig:sic_monthly_result} shows the monthly averaged MAE and RMSE, along with their range deviated by one standard deviation of Unicorn, and compared models in the test set. 
As depicted in Figure \ref{fig:sic_monthly_result}, during the sea ice decreasing overall period, Unicorn achieves superior performance over the state-of-the-art models in SIC forecasting. This indicates that Unicorn can effectively capture spatiotemporal patterns of SIC better than others. 

Figure \ref{fig:sic_result} displays the 4-week ahead SIC forecasting results of selected models: CNN, U-Net, SICNet (CBAM), and Unicorn. The red color represents the absolute error between prediction and ground truth. At the top of the images, the MAE value is presented alongside the model's name. Unicorn is able to capture an increasing trend in the edge region of the sea ice, whereas the other models fail to do so as the target time point increases.

\subsection{SIE forecasting performance}

\begin{figure}[t]
    \centering
    \subfigure{\includegraphics[width=0.9\linewidth]{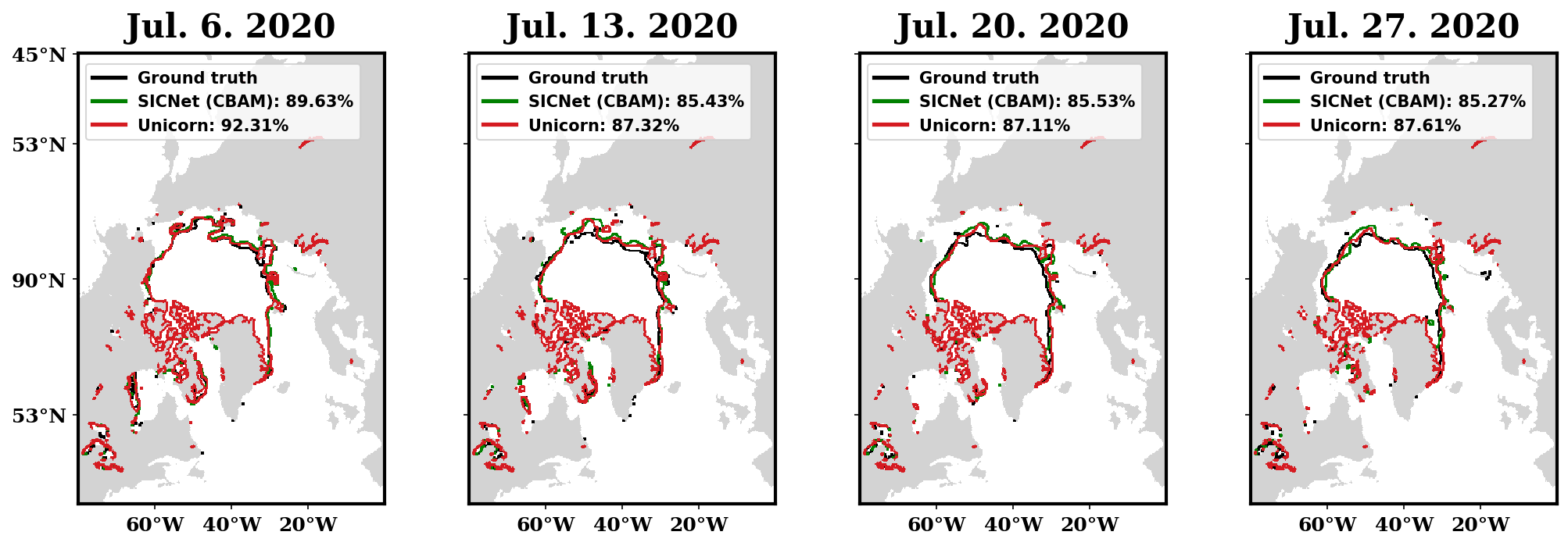}}
    \subfigure{\includegraphics[width=0.9\linewidth]{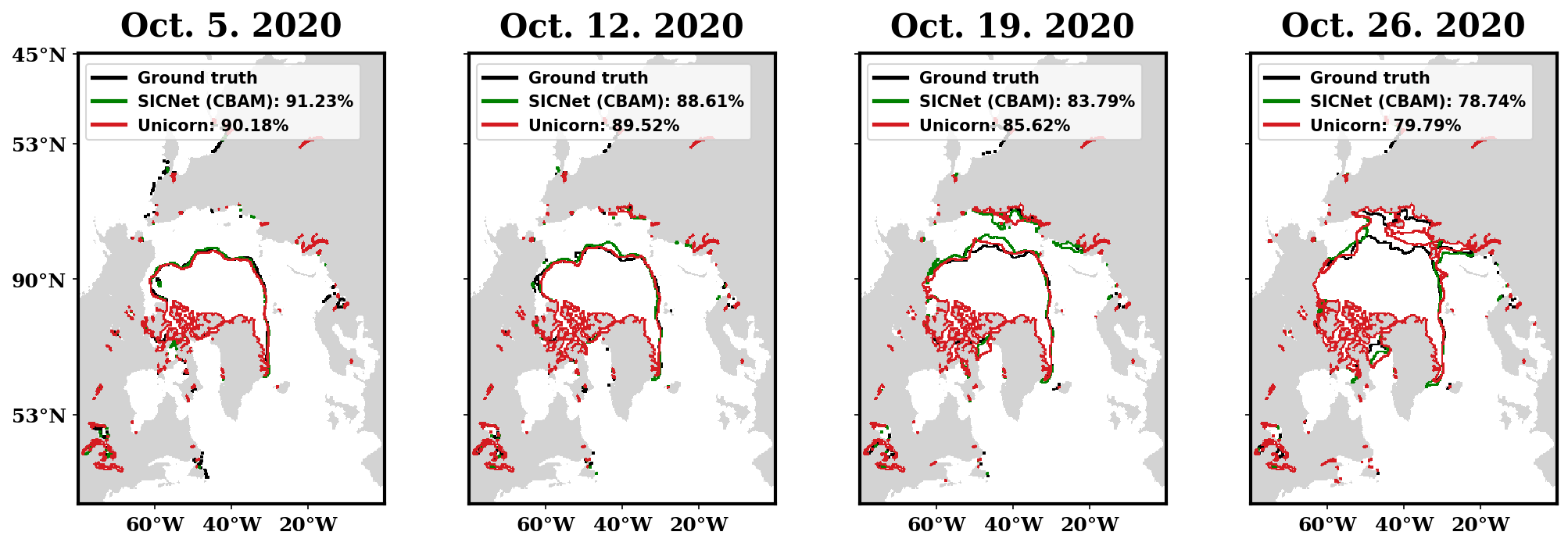}}
    \caption{SIE forecasting results. In the image, the boundaries of SIE forecasting for SICNet (CBAM), Unicorn, and the ground truth are shown along with their mIoU values.}
    \label{fig:sie_result}
\end{figure}

\begin{figure}[]
    \centering
    \includegraphics[width=0.7\linewidth]{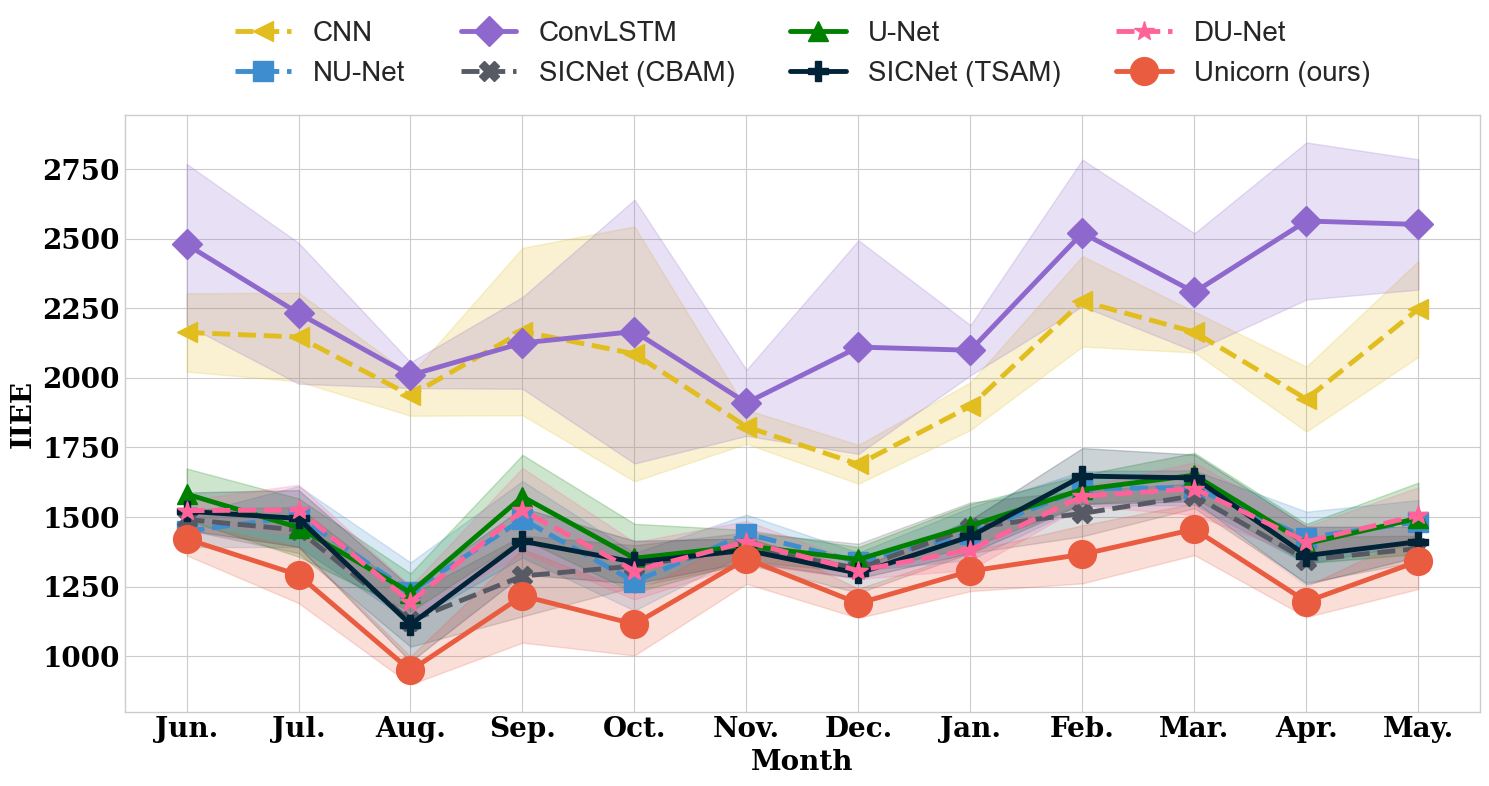}
    \caption{Monthly average IIEE forecasting performance of Unicorn and compared models from June 2020 to May 2021.
    }
    \label{fig:sie_monthly_result}
\end{figure}

In the sea ice domain, accurate SIE forecasting is equally as important as SIC forecasting. Table \ref{tab:results} presents the performance of SIE forecasting with three metrics: IIEE, mIoU, and F1-score. In all three metrics, Unicorn outperforms the state-of-the-art models. The Unicorn model exhibits enhancements in SIE forecasting, particularly in the IIEE metric, where it shows an average improvement of 17.59\% with a 7.55$\%$ improvement over its closest competitor, SICNet (CBAM). The mIoU and F1 scores still signify advancements over existing models. These improvements underscore Unicorn's consistent accuracy and reliability across varying ice conditions.

Figure \ref{fig:bar_charts} (bottom) shows the average IIEE for different target time points.
As with the comparison of the average RMSE, there are significant performance differences between the U-Net-based models and others.
Notably, Unicorn has the lowest IIEE at each target time point compared to other models. 
Figure \ref{fig:sie_result} displays the 4-week ahead SIE forecasting results of SICNet (CBAM), the next best model, and Unicorn. The first row of Figure \ref{fig:sie_result} presents the results from June 7, 2020, to July 27, 2020, during a period when the sea ice exhibits a decreasing pattern as shown in Figure \ref{fig:sic_result}. The second row of Figure \ref{fig:sie_result} presents the results from October 5, 2020, to October 26, 2020, during a period when the sea ice exhibits an increasing pattern. The black line represents the ground truth sea ice edge, while the green and red lines represent the sea ice edge based on SICNet (CBAM) and Unicorn results, respectively. At the top of the images, the mIoU values between the ground truth and prediction model are displayed alongside the model's name. In both cases (decreasing and increasing periods), Unicorn outperforms SICNet (CBAM). 

Figure \ref{fig:sie_monthly_result} shows the monthly averaged IIEE, along with their range deviated by one standard deviation of Unicorn, and compared models in the test set. 
As shown in Figure \ref{fig:sie_monthly_result} during the overall period, Unicorn achieves outstanding performance compared to other SIE forecasting models. 

\subsection{Ablation study} \label{sec:abl} 

In this study, we performed an ablation analysis of the Unicorn model to evaluate the contributions of various components -- DCMP, ConvNODE, and ancillary datasets -- to the accuracy of SIC and SIE forecasting. By systematically removing each component, we can isolate and assess their individual impacts. The findings of this analysis are detailed in Table \ref{tab:ablation}. This structured approach allows us to clearly identify the role of each component in improving the model's performance.

\begin{table}[b]
\caption{Experimental results of the ablation study. The most favorable value is highlighted in bold.}
\centering
\begin{tabular}{lcccccc}
\toprule            
 task & \multicolumn{2}{c}{SIC} & \multicolumn{3}{c}{SIE}\\
 \cmidrule(lr){2-3} \cmidrule(lr){4-6}
 Metric & MAE $\downarrow$ & RMSE $\downarrow$ & IIEE $\downarrow$ & mIoU $\uparrow$& F1 $\uparrow$ \\ 
\midrule
U-Net & 0.0245 & 0.0769 & 1428.0643 & 0.9107 & 0.9528 \\ 
Unicorn w/o DCMP & 0.0231 & 0.0713 & 1299.8065 & 0.9189 & 0.9574 \\
Unicorn w/o ConvNODE & 0.0231 & 0.0726 & 1302.4603 & 0.9181 & 0.9570 \\
Unicorn w/o ancillary data & 0.0238 & 0.0733 & 1376.9588 & 0.9141 & 0.9547 \\
Unicorn & \textbf{0.0228} & \textbf{0.0707} & \textbf{1270.4770} & \textbf{0.9204} & \textbf{0.9583} \\ 
\botrule
\end{tabular}
\label{tab:ablation}
\end{table}

\begin{figure}[t]
    \centering
    \includegraphics[width=0.9\linewidth]{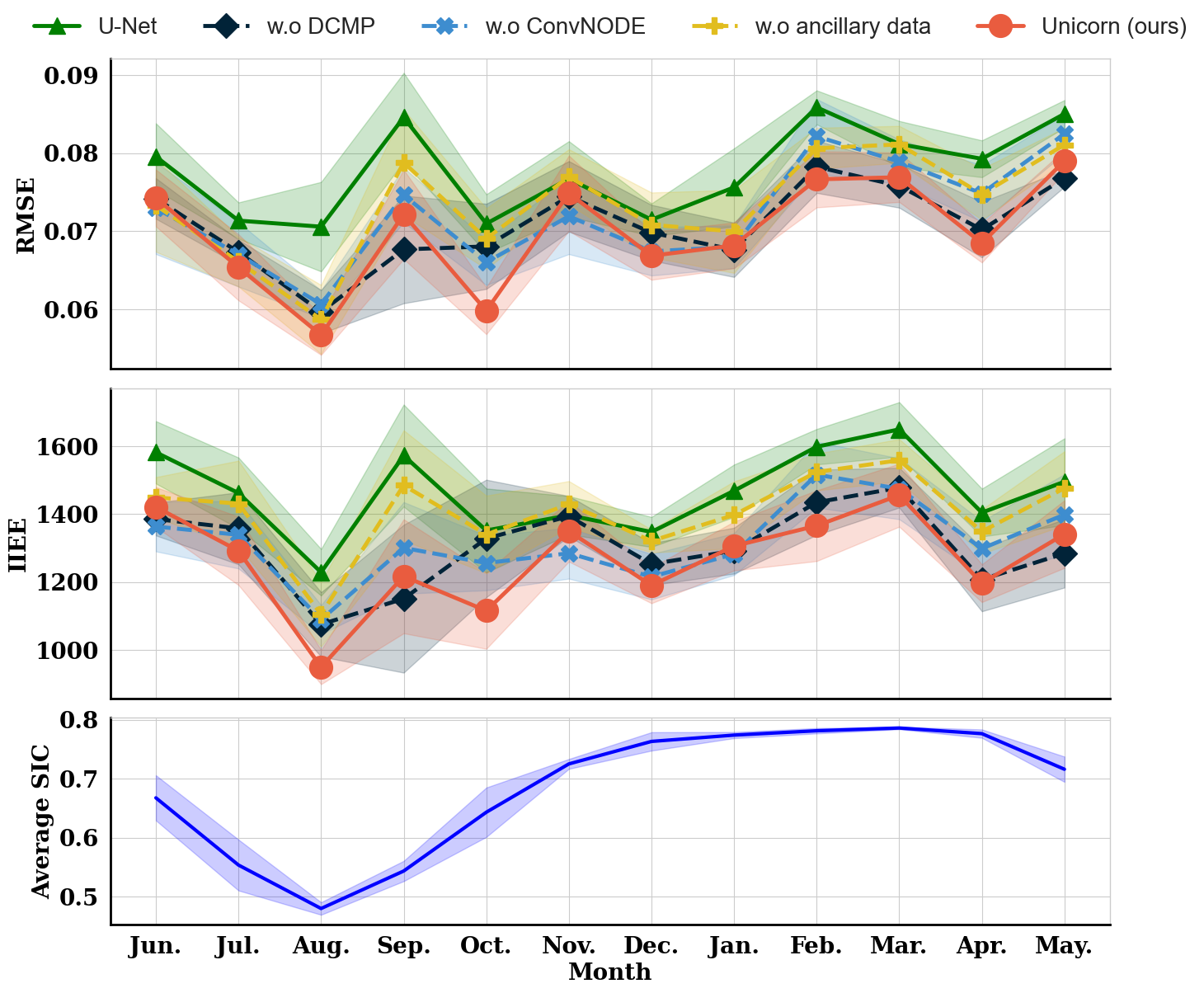}
    \caption{Monthly average RMSE (top), IIEE (middle), and SIC (bottom) of Unicorn and ablated models. The interval presents the average $\pm 1$ standard error.}
    \label{fig:abl_monthly}
\end{figure}

Our ablation analysis reveals significant insights into the contributions of each component within the Unicorn model. 
The removal of ancillary data had the most substantial impact, causing significant declines in both SIC and SIE predictions. Specifically, MAE increased by 4.20\%, RMSE by 3.55\%, IIEE by 7.73\%, mIoU by 0.69\%, and the F1 score by 0.38\%. 
As discussed in Section \ref{sec:data}, TB and SIA datasets play a crucial role in supporting SIC forecasting by providing additional information, such as edge segmentation based on sea ice ages. 
The omission of ConvNODE from Unicorn resulted in notable accuracy losses, specifically a 1.30\% increase in MAE and a 2.62\% increase in RMSE for SIC forecasting. This demonstrates the essential role of ConvNODE in maintaining forecasting accuracy.

In contrast, removing the DCMP block led to a 2.26\% decrease in IIEE for SIE forecasting. To delve into the efficiency of the DCMP block, we made a minute comparison of performances. Figure \ref{fig:abl_monthly} displays the monthly average performance and SIC for highlighting forecasting performances in non-stationary scenarios. Unicorn shows the overall best performance in both metrics. However, Unicorn without the DCMP block still shows a competitive performance.
It is worth noting that when there are decreasing and increasing SIC patterns in order, from June to October, the Unicorn outperforms its ablated model without the DCMP block. This highlights the effectiveness of DCMP under the distribution shift scenario. 

These above results, through rigorous cross-validation, highlight the essential role of each component in ensuring the model's effectiveness across different conditions. The ablation study confirms that Unicorn performs optimally when all components are fully integrated, consistently surpassing the baseline model, U-Net, even when individual components are removed.

\section{Conclusion} \label{sec:conc}
Our study presents a novel approach to sea ice forecasting from two perspectives. Firstly, we developed an enhanced ConvNODE model through local time series decomposition to account for the non-stationarity of sea ice dynamics. Secondly, we propose integrating additional information from brightness temperature and sea ice age.
Through real data analysis from 1998 to 2021, Unicorn shows improvements, with a 12\% enhancement in the average MAE for SIC and approximately an 18\% improvement in the primary SIE metric IIEE. This advancement is expected to greatly enhance the precision of sea ice forecasts and stimulate further research in climate and related fields.

Although the proposed model achieves high-quality performance, it still has some limitations. One notable limitation is its high computing cost. Our model generates latent variables and images in multiple timestamps through respective upsampling paths, resulting in a relatively higher number of trainable parameters than other models. Another limitation is the absence of explainability. Due to the nature of neural networks, the outputs of the proposed model are challenging to explain. We propose incorporating our model with an interpretable neural network module such as the variable selection network \cite{lim2021temporal} as a potential solution. Another approach is to define the sea ice forecasting problem as a multi-task problem, encompassing both SIC and SIE forecasting, as demonstrated in prior work \cite{Ali2022MTIceNetA}. In this setup, canonical correlation analysis can be employed to extract canonical information from multiple tasks \cite{rakowski2023dcid}, aiding in learning shared and task-specific representations and providing explanations for forecasting results. We consider addressing these issues as part of our future work.

\bibliography{sic}


\begin{thebibliography}{50}
\ifx \bisbn   \undefined \def \bisbn  #1{ISBN #1}\fi
\ifx \binits  \undefined \def \binits#1{#1}\fi
\ifx \bauthor  \undefined \def \bauthor#1{#1}\fi
\ifx \batitle  \undefined \def \batitle#1{#1}\fi
\ifx \bjtitle  \undefined \def \bjtitle#1{#1}\fi
\ifx \bvolume  \undefined \def \bvolume#1{\textbf{#1}}\fi
\ifx \byear  \undefined \def \byear#1{#1}\fi
\ifx \bissue  \undefined \def \bissue#1{#1}\fi
\ifx \bfpage  \undefined \def \bfpage#1{#1}\fi
\ifx \blpage  \undefined \def \blpage #1{#1}\fi
\ifx \burl  \undefined \def \burl#1{\textsf{#1}}\fi
\ifx \doiurl  \undefined \def \doiurl#1{\url{https://doi.org/#1}}\fi
\ifx \betal  \undefined \def \betal{\textit{et al.}}\fi
\ifx \binstitute  \undefined \def \binstitute#1{#1}\fi
\ifx \binstitutionaled  \undefined \def \binstitutionaled#1{#1}\fi
\ifx \bctitle  \undefined \def \bctitle#1{#1}\fi
\ifx \beditor  \undefined \def \beditor#1{#1}\fi
\ifx \bpublisher  \undefined \def \bpublisher#1{#1}\fi
\ifx \bbtitle  \undefined \def \bbtitle#1{#1}\fi
\ifx \bedition  \undefined \def \bedition#1{#1}\fi
\ifx \bseriesno  \undefined \def \bseriesno#1{#1}\fi
\ifx \blocation  \undefined \def \blocation#1{#1}\fi
\ifx \bsertitle  \undefined \def \bsertitle#1{#1}\fi
\ifx \bsnm \undefined \def \bsnm#1{#1}\fi
\ifx \bsuffix \undefined \def \bsuffix#1{#1}\fi
\ifx \bparticle \undefined \def \bparticle#1{#1}\fi
\ifx \barticle \undefined \def \barticle#1{#1}\fi
\bibcommenthead
\ifx \bconfdate \undefined \def \bconfdate #1{#1}\fi
\ifx \botherref \undefined \def \botherref #1{#1}\fi
\ifx \url \undefined \def \url#1{\textsf{#1}}\fi
\ifx \bchapter \undefined \def \bchapter#1{#1}\fi
\ifx \bbook \undefined \def \bbook#1{#1}\fi
\ifx \bcomment \undefined \def \bcomment#1{#1}\fi
\ifx \oauthor \undefined \def \oauthor#1{#1}\fi
\ifx \citeauthoryear \undefined \def \citeauthoryear#1{#1}\fi
\ifx \endbibitem  \undefined \def \endbibitem {}\fi
\ifx \bconflocation  \undefined \def \bconflocation#1{#1}\fi
\ifx \arxivurl  \undefined \def \arxivurl#1{\textsf{#1}}\fi
\csname PreBibitemsHook\endcsname

\bibitem[\protect\citeauthoryear{Notz and Stroeve}{2016}]{Notz2016ObservedAS}
\begin{barticle}
\bauthor{\bsnm{Notz}, \binits{D.}},
\bauthor{\bsnm{Stroeve}, \binits{J.C.}}:
\batitle{Observed arctic sea-ice loss directly follows anthropogenic co2 emission}.
\bjtitle{Science}
\bvolume{354},
\bfpage{747}--\blpage{750}
(\byear{2016})
\end{barticle}
\endbibitem

\bibitem[\protect\citeauthoryear{Wunderling et~al.}{2020}]{Wunderling2020GlobalWD}
\begin{botherref}
\oauthor{\bsnm{Wunderling}, \binits{N.}},
\oauthor{\bsnm{Willeit}, \binits{M.}},
\oauthor{\bsnm{Donges}, \binits{J.F.}},
\oauthor{\bsnm{Winkelmann}, \binits{R.}}:
Global warming due to loss of large ice masses and arctic summer sea ice.
Nature Communications
\textbf{11}
(2020)
\end{botherref}
\endbibitem

\bibitem[\protect\citeauthoryear{Rantanen et~al.}{2022}]{Rantanen2022TheAH}
\begin{botherref}
\oauthor{\bsnm{Rantanen}, \binits{M.}},
\oauthor{\bsnm{Karpechko}, \binits{A.Y.}},
\oauthor{\bsnm{Lipponen}, \binits{A.}},
\oauthor{\bsnm{Nordling}, \binits{K.}},
\oauthor{\bsnm{Hyv{\"a}rinen}, \binits{O.}},
\oauthor{\bsnm{Ruosteenoja}, \binits{K.}},
\oauthor{\bsnm{Vihma}, \binits{T.}},
\oauthor{\bsnm{Laaksonen}, \binits{A.}}:
The arctic has warmed nearly four times faster than the globe since 1979.
Communications Earth \& Environment
\textbf{3}
(2022)
\end{botherref}
\endbibitem

\bibitem[\protect\citeauthoryear{Screen and Simmonds}{2010}]{Screen2010TheCR}
\begin{barticle}
\bauthor{\bsnm{Screen}, \binits{J.A.}},
\bauthor{\bsnm{Simmonds}, \binits{I.}}:
\batitle{The central role of diminishing sea ice in recent arctic temperature amplification}.
\bjtitle{Nature}
\bvolume{464},
\bfpage{1334}--\blpage{1337}
(\byear{2010})
\end{barticle}
\endbibitem

\bibitem[\protect\citeauthoryear{Pithan and Mauritsen}{2014}]{Pithan2014ArcticAD}
\begin{barticle}
\bauthor{\bsnm{Pithan}, \binits{F.}},
\bauthor{\bsnm{Mauritsen}, \binits{T.}}:
\batitle{Arctic amplification dominated by temperature feedbacks in contemporary climate models}.
\bjtitle{Nature Geoscience}
\bvolume{7},
\bfpage{181}--\blpage{184}
(\byear{2014})
\end{barticle}
\endbibitem

\bibitem[\protect\citeauthoryear{Li et~al.}{2009}]{Li2009SmallestAT}
\begin{barticle}
\bauthor{\bsnm{Li}, \binits{W.K.W.}},
\bauthor{\bsnm{McLaughlin}, \binits{F.A.}},
\bauthor{\bsnm{Lovejoy}, \binits{C.}},
\bauthor{\bsnm{Carmack}, \binits{E.C.}}:
\batitle{Smallest algae thrive as the arctic ocean freshens}.
\bjtitle{Science}
\bvolume{326},
\bfpage{539}--\blpage{539}
(\byear{2009})
\end{barticle}
\endbibitem

\bibitem[\protect\citeauthoryear{Kraemer et~al.}{2024}]{Kraemer2024AMT}
\begin{botherref}
\oauthor{\bsnm{Kraemer}, \binits{S.A.}},
\oauthor{\bsnm{Ramachandran}, \binits{A.}},
\oauthor{\bsnm{Onana}, \binits{V.E.}},
\oauthor{\bsnm{Li}, \binits{W.K.W.}},
\oauthor{\bsnm{Walsh}, \binits{D.A.}}:
A multiyear time series (2004–2012) of bacterial and archaeal community dynamics in a changing arctic ocean.
ISME Communications
\textbf{4}
(2024)
\end{botherref}
\endbibitem

\bibitem[\protect\citeauthoryear{Wei et~al.}{2022}]{Wei2022PredictionOP}
\begin{bchapter}
\bauthor{\bsnm{Wei}, \binits{J.}},
\bauthor{\bsnm{Hang}, \binits{R.}},
\bauthor{\bsnm{Luo}, \binits{J.}}:
\bctitle{Prediction of pan-arctic sea ice using attention-based lstm neural networks}.
In: \bbtitle{Frontiers in Marine Science}
(\byear{2022})
\end{bchapter}
\endbibitem

\bibitem[\protect\citeauthoryear{Kwok}{2018}]{Kwok2018ArcticSI}
\begin{botherref}
\oauthor{\bsnm{Kwok}, \binits{R.}}:
Arctic sea ice thickness, volume, and multiyear ice coverage: losses and coupled variability (1958–2018).
Environmental Research Letters
\textbf{13}
(2018)
\end{botherref}
\endbibitem

\bibitem[\protect\citeauthoryear{Sumata et~al.}{2023}]{Sumata2023RegimeSI}
\begin{barticle}
\bauthor{\bsnm{Sumata}, \binits{H.}},
\bauthor{\bsnm{Steur}, \binits{L.}},
\bauthor{\bsnm{Divine}, \binits{D.V.}},
\bauthor{\bsnm{Granskog}, \binits{M.A.}},
\bauthor{\bsnm{Gerland}, \binits{S.}}:
\batitle{Regime shift in arctic ocean sea ice thickness}.
\bjtitle{Nature}
\bvolume{615},
\bfpage{443}--\blpage{449}
(\byear{2023})
\end{barticle}
\endbibitem

\bibitem[\protect\citeauthoryear{Notz and Community}{2020}]{Notz2020ArcticSI}
\begin{botherref}
\oauthor{\bsnm{Notz}, \binits{D.}},
\oauthor{\bsnm{Community}, \binits{S.}}:
Arctic sea ice in cmip6.
Geophysical Research Letters
\textbf{47}
(2020)
\end{botherref}
\endbibitem

\bibitem[\protect\citeauthoryear{Kim et~al.}{2023}]{Kim2023ObservationallyconstrainedPO}
\begin{barticle}
\bauthor{\bsnm{Kim}, \binits{Y.-H.}},
\bauthor{\bsnm{Min}, \binits{S.-K.}},
\bauthor{\bsnm{Gillett}, \binits{N.P.}},
\bauthor{\bsnm{Notz}, \binits{D.}},
\bauthor{\bsnm{Malinina}, \binits{E.}}:
\batitle{Observationally-constrained projections of an ice-free arctic even under a low emission scenario}.
\bjtitle{Nature Communications}
\bvolume{14}(\bissue{1}),
\bfpage{3139}
(\byear{2023})
\end{barticle}
\endbibitem

\bibitem[\protect\citeauthoryear{Wang et~al.}{2013}]{Wang2013SeasonalPO}
\begin{barticle}
\bauthor{\bsnm{Wang}, \binits{W.}},
\bauthor{\bsnm{Chen}, \binits{M.}},
\bauthor{\bsnm{Kumar}, \binits{A.}}:
\batitle{Seasonal prediction of arctic sea ice extent from a coupled dynamical forecast system}.
\bjtitle{Monthly Weather Review}
\bvolume{141},
\bfpage{1375}--\blpage{1394}
(\byear{2013})
\end{barticle}
\endbibitem

\bibitem[\protect\citeauthoryear{Collow et~al.}{2015}]{collow2015improving}
\begin{barticle}
\bauthor{\bsnm{Collow}, \binits{T.W.}},
\bauthor{\bsnm{Wang}, \binits{W.}},
\bauthor{\bsnm{Kumar}, \binits{A.}},
\bauthor{\bsnm{Zhang}, \binits{J.}}:
\batitle{Improving arctic sea ice prediction using piomas initial sea ice thickness in a coupled ocean--atmosphere model}.
\bjtitle{Monthly Weather Review}
\bvolume{143}(\bissue{11}),
\bfpage{4618}--\blpage{4630}
(\byear{2015})
\end{barticle}
\endbibitem

\bibitem[\protect\citeauthoryear{Yuan et~al.}{2016}]{yuan2016arctic}
\begin{barticle}
\bauthor{\bsnm{Yuan}, \binits{X.}},
\bauthor{\bsnm{Chen}, \binits{D.}},
\bauthor{\bsnm{Li}, \binits{C.}},
\bauthor{\bsnm{Wang}, \binits{L.}},
\bauthor{\bsnm{Wang}, \binits{W.}}:
\batitle{Arctic sea ice seasonal prediction by a linear markov model}.
\bjtitle{Journal of Climate}
\bvolume{29}(\bissue{22}),
\bfpage{8151}--\blpage{8173}
(\byear{2016})
\end{barticle}
\endbibitem

\bibitem[\protect\citeauthoryear{Choi et~al.}{2019}]{Choi2019ArtificialNN}
\begin{barticle}
\bauthor{\bsnm{Choi}, \binits{M.}},
\bauthor{\bsnm{Silva}, \binits{L.W.A.D.}},
\bauthor{\bsnm{Yamaguchi}, \binits{H.}}:
\batitle{Artificial neural network for the short-term prediction of arctic sea ice concentration}.
\bjtitle{Remote. Sens.}
\bvolume{11},
\bfpage{1071}
(\byear{2019})
\end{barticle}
\endbibitem

\bibitem[\protect\citeauthoryear{He et~al.}{2022}]{He2022AnIC}
\begin{bchapter}
\bauthor{\bsnm{He}, \binits{J.}},
\bauthor{\bsnm{Zhao}, \binits{Y.}},
\bauthor{\bsnm{Yang}, \binits{D.}},
\bauthor{\bsnm{Zhu}, \binits{K.}},
\bauthor{\bsnm{Deng}, \binits{X.}}:
\bctitle{An improved convlstm network for arctic sea ice concentration prediction}.
In: \bbtitle{OCEANS 2022, Hampton Roads},
pp. \bfpage{1}--\blpage{5}
(\byear{2022}).
\bcomment{IEEE}
\end{bchapter}
\endbibitem

\bibitem[\protect\citeauthoryear{Kim et~al.}{2023}]{kim2023polargan}
\begin{barticle}
\bauthor{\bsnm{Kim}, \binits{M.}},
\bauthor{\bsnm{Lee}, \binits{J.}},
\bauthor{\bsnm{Choi}, \binits{L.}},
\bauthor{\bsnm{Choi}, \binits{M.}}:
\batitle{Polargan: Creating realistic arctic sea ice concentration images with user-defined geometric preferences}.
\bjtitle{Engineering Applications of Artificial Intelligence}
\bvolume{126},
\bfpage{106920}
(\byear{2023})
\end{barticle}
\endbibitem

\bibitem[\protect\citeauthoryear{Andersson et~al.}{2021}]{Andersson2021SeasonalAS}
\begin{barticle}
\bauthor{\bsnm{Andersson}, \binits{T.R.}},
\bauthor{\bsnm{Hosking}, \binits{J.S.}},
\bauthor{\bsnm{P{\'e}rez-Ortiz}, \binits{M.}},
\bauthor{\bsnm{Paige}, \binits{B.}},
\bauthor{\bsnm{Elliott}, \binits{A.}},
\bauthor{\bsnm{Russell}, \binits{C.}},
\bauthor{\bsnm{Law}, \binits{S.}},
\bauthor{\bsnm{Jones}, \binits{D.C.}},
\bauthor{\bsnm{Wilkinson}, \binits{J.}},
\bauthor{\bsnm{Phillips}, \binits{T.}}, \betal:
\batitle{Seasonal arctic sea ice forecasting with probabilistic deep learning}.
\bjtitle{Nature communications}
\bvolume{12}(\bissue{1}),
\bfpage{5124}
(\byear{2021})
\end{barticle}
\endbibitem

\bibitem[\protect\citeauthoryear{Ren et~al.}{2022}]{ren2022data}
\begin{barticle}
\bauthor{\bsnm{Ren}, \binits{Y.}},
\bauthor{\bsnm{Li}, \binits{X.}},
\bauthor{\bsnm{Zhang}, \binits{W.}}:
\batitle{A data-driven deep learning model for weekly sea ice concentration prediction of the pan-arctic during the melting season}.
\bjtitle{IEEE Transactions on Geoscience and Remote Sensing}
\bvolume{60},
\bfpage{1}--\blpage{19}
(\byear{2022})
\end{barticle}
\endbibitem

\bibitem[\protect\citeauthoryear{Chen et~al.}{2018}]{chen2018neural}
\begin{bchapter}
\bauthor{\bsnm{Chen}, \binits{R.T.}},
\bauthor{\bsnm{Rubanova}, \binits{Y.}},
\bauthor{\bsnm{Bettencourt}, \binits{J.}},
\bauthor{\bsnm{Duvenaud}, \binits{D.}}:
\bctitle{Neural ordinary differential equations}.
In: \bbtitle{Proceedings of the 32nd International Conference on Neural Information Processing Systems},
pp. \bfpage{6572}--\blpage{6583}
(\byear{2018})
\end{bchapter}
\endbibitem

\bibitem[\protect\citeauthoryear{Paoletti et~al.}{2019}]{paoletti2019neural}
\begin{barticle}
\bauthor{\bsnm{Paoletti}, \binits{M.E.}},
\bauthor{\bsnm{Haut}, \binits{J.M.}},
\bauthor{\bsnm{Plaza}, \binits{J.}},
\bauthor{\bsnm{Plaza}, \binits{A.}}:
\batitle{Neural ordinary differential equations for hyperspectral image classification}.
\bjtitle{IEEE Transactions on Geoscience and Remote Sensing}
\bvolume{58}(\bissue{3}),
\bfpage{1718}--\blpage{1734}
(\byear{2019})
\end{barticle}
\endbibitem

\bibitem[\protect\citeauthoryear{Li et~al.}{2021}]{li2021robust}
\begin{botherref}
\oauthor{\bsnm{Li}, \binits{D.}},
\oauthor{\bsnm{Tang}, \binits{P.}},
\oauthor{\bsnm{Zhang}, \binits{R.}},
\oauthor{\bsnm{Sun}, \binits{C.}},
\oauthor{\bsnm{Li}, \binits{Y.}},
\oauthor{\bsnm{Qian}, \binits{J.}},
\oauthor{\bsnm{Liang}, \binits{Y.}},
\oauthor{\bsnm{Yang}, \binits{J.}},
\oauthor{\bsnm{Zhang}, \binits{L.}}:
Robust blood cell image segmentation method based on neural ordinary differential equations.
Computational and Mathematical Methods in Medicine
\textbf{2021}
(2021)
\end{botherref}
\endbibitem

\bibitem[\protect\citeauthoryear{Salles et~al.}{2019}]{salles2019nonstationary}
\begin{barticle}
\bauthor{\bsnm{Salles}, \binits{R.}},
\bauthor{\bsnm{Belloze}, \binits{K.}},
\bauthor{\bsnm{Porto}, \binits{F.}},
\bauthor{\bsnm{Gonzalez}, \binits{P.H.}},
\bauthor{\bsnm{Ogasawara}, \binits{E.}}:
\batitle{Nonstationary time series transformation methods: An experimental review}.
\bjtitle{Knowledge-Based Systems}
\bvolume{164},
\bfpage{274}--\blpage{291}
(\byear{2019})
\end{barticle}
\endbibitem

\bibitem[\protect\citeauthoryear{Wang et~al.}{2016}]{Wang2016PredictingSA}
\begin{barticle}
\bauthor{\bsnm{Wang}, \binits{L.}},
\bauthor{\bsnm{Yuan}, \binits{X.}},
\bauthor{\bsnm{Ting}, \binits{M.}},
\bauthor{\bsnm{Li}, \binits{C.}}:
\batitle{Predicting summer arctic sea ice concentration intraseasonal variability using a vector autoregressive model}.
\bjtitle{Journal of Climate}
\bvolume{29},
\bfpage{1529}--\blpage{1543}
(\byear{2016})
\end{barticle}
\endbibitem

\bibitem[\protect\citeauthoryear{Chi and Kim}{2017}]{Chi2017PredictionOA}
\begin{barticle}
\bauthor{\bsnm{Chi}, \binits{J.}},
\bauthor{\bsnm{Kim}, \binits{H.}}:
\batitle{Prediction of arctic sea ice concentration using a fully data driven deep neural network}.
\bjtitle{Remote. Sens.}
\bvolume{9},
\bfpage{1305}
(\byear{2017})
\end{barticle}
\endbibitem

\bibitem[\protect\citeauthoryear{Song et~al.}{2018}]{Song2018ARC}
\begin{botherref}
\oauthor{\bsnm{Song}, \binits{W.}},
\oauthor{\bsnm{Li}, \binits{M.}},
\oauthor{\bsnm{He}, \binits{Q.}},
\oauthor{\bsnm{Huang}, \binits{D.}},
\oauthor{\bsnm{Perra}, \binits{C.}},
\oauthor{\bsnm{Liotta}, \binits{A.}}:
A residual convolution neural network for sea ice classification with sentinel-1 sar imagery.
2018 IEEE International Conference on Data Mining Workshops (ICDMW),
795--802
(2018)
\end{botherref}
\endbibitem

\bibitem[\protect\citeauthoryear{Shi et~al.}{2015}]{shi2015convolutional}
\begin{botherref}
\oauthor{\bsnm{Shi}, \binits{X.}},
\oauthor{\bsnm{Chen}, \binits{Z.}},
\oauthor{\bsnm{Wang}, \binits{H.}},
\oauthor{\bsnm{Yeung}, \binits{D.-Y.}},
\oauthor{\bsnm{Wong}, \binits{W.-K.}},
\oauthor{\bsnm{Woo}, \binits{W.-c.}}:
Convolutional lstm network: A machine learning approach for precipitation nowcasting.
Advances in neural information processing systems
\textbf{28}
(2015)
\end{botherref}
\endbibitem

\bibitem[\protect\citeauthoryear{Ronneberger et~al.}{2015}]{ronneberger2015u}
\begin{bchapter}
\bauthor{\bsnm{Ronneberger}, \binits{O.}},
\bauthor{\bsnm{Fischer}, \binits{P.}},
\bauthor{\bsnm{Brox}, \binits{T.}}:
\bctitle{U-net: Convolutional networks for biomedical image segmentation}.
In: \bbtitle{Medical Image Computing and Computer-Assisted Intervention--MICCAI 2015: 18th International Conference, Munich, Germany, October 5-9, 2015, Proceedings, Part III 18},
pp. \bfpage{234}--\blpage{241}
(\byear{2015}).
\bcomment{Springer}
\end{bchapter}
\endbibitem

\bibitem[\protect\citeauthoryear{Huang et~al.}{2023}]{huang2023interpretable}
\begin{barticle}
\bauthor{\bsnm{Huang}, \binits{X.}},
\bauthor{\bsnm{Zhang}, \binits{B.}},
\bauthor{\bsnm{Feng}, \binits{S.}},
\bauthor{\bsnm{Ye}, \binits{Y.}},
\bauthor{\bsnm{Li}, \binits{X.}}:
\batitle{Interpretable local flow attention for multi-step traffic flow prediction}.
\bjtitle{Neural networks}
\bvolume{161},
\bfpage{25}--\blpage{38}
(\byear{2023})
\end{barticle}
\endbibitem

\bibitem[\protect\citeauthoryear{Kim et~al.}{2019}]{kim2019learning}
\begin{bchapter}
\bauthor{\bsnm{Kim}, \binits{S.}},
\bauthor{\bsnm{Park}, \binits{S.}},
\bauthor{\bsnm{Chung}, \binits{S.}},
\bauthor{\bsnm{Lee}, \binits{J.}},
\bauthor{\bsnm{Lee}, \binits{Y.}},
\bauthor{\bsnm{Kim}, \binits{H.}},
\bauthor{\bsnm{Prabhat}, \binits{M.}},
\bauthor{\bsnm{Choo}, \binits{J.}}:
\bctitle{Learning to focus and track extreme climate events}.
In: \bbtitle{30th British Machine Vision Conference, {BMVC} 2019, Cardiff, UK, September 9-12, 2019},
p. \bfpage{11}.
\bpublisher{{BMVA} Press}, \blocation{???}
(\byear{2019})
\end{bchapter}
\endbibitem

\bibitem[\protect\citeauthoryear{Liu et~al.}{2021}]{Liu2021DailyPO}
\begin{barticle}
\bauthor{\bsnm{Liu}, \binits{Q.}},
\bauthor{\bsnm{Zhang}, \binits{R.}},
\bauthor{\bsnm{Wang}, \binits{Y.}},
\bauthor{\bsnm{Yan}, \binits{H.}},
\bauthor{\bsnm{Hong}, \binits{M.}}:
\batitle{Daily prediction of the arctic sea ice concentration using reanalysis data based on a convolutional lstm network}.
\bjtitle{Journal of Marine Science and Engineering}
\bvolume{9},
\bfpage{330}
(\byear{2021})
\end{barticle}
\endbibitem

\bibitem[\protect\citeauthoryear{Kim et~al.}{2021}]{Kim2021MultiTaskDL}
\begin{bchapter}
\bauthor{\bsnm{Kim}, \binits{E.}},
\bauthor{\bsnm{Kruse}, \binits{P.}},
\bauthor{\bsnm{Lama}, \binits{S.}},
\bauthor{\bsnm{Bourne}, \binits{J.}},
\bauthor{\bsnm{Hu}, \binits{M.}},
\bauthor{\bsnm{Ali}, \binits{S.}},
\bauthor{\bsnm{Huang}, \binits{Y.}},
\bauthor{\bsnm{Wang}, \binits{J.}}:
\bctitle{Multi-task deep learning based spatiotemporal arctic sea ice forecasting}.
In: \bbtitle{2021 IEEE International Conference on Big Data (Big Data)},
pp. \bfpage{1847}--\blpage{1857}
(\byear{2021}).
\bcomment{IEEE}
\end{bchapter}
\endbibitem

\bibitem[\protect\citeauthoryear{Seo et~al.}{2022}]{seo2022simple}
\begin{botherref}
\oauthor{\bsnm{Seo}, \binits{M.}},
\oauthor{\bsnm{Kim}, \binits{D.}},
\oauthor{\bsnm{Shin}, \binits{S.}},
\oauthor{\bsnm{Kim}, \binits{E.}},
\oauthor{\bsnm{Ahn}, \binits{S.}},
\oauthor{\bsnm{Choi}, \binits{Y.}}:
Simple baseline for weather forecasting using spatiotemporal context aggregation network.
arXiv preprint arXiv:2212.02952
(2022)
\end{botherref}
\endbibitem

\bibitem[\protect\citeauthoryear{Fern{\'a}ndez and Mehrkanoon}{2021}]{fernandez2021broad}
\begin{barticle}
\bauthor{\bsnm{Fern{\'a}ndez}, \binits{J.G.}},
\bauthor{\bsnm{Mehrkanoon}, \binits{S.}}:
\batitle{Broad-unet: Multi-scale feature learning for nowcasting tasks}.
\bjtitle{Neural Networks}
\bvolume{144},
\bfpage{419}--\blpage{427}
(\byear{2021})
\end{barticle}
\endbibitem

\bibitem[\protect\citeauthoryear{Choi}{2020}]{choi2020utilizing}
\begin{botherref}
\oauthor{\bsnm{Choi}, \binits{S.}}:
Utilizing unet for the future traffic map prediction task traffic4cast challenge 2020.
arXiv preprint arXiv:2012.00125
(2020)
\end{botherref}
\endbibitem

\bibitem[\protect\citeauthoryear{Ivanova et~al.}{2015}]{Ivanova2015IntercomparisonAE}
\begin{barticle}
\bauthor{\bsnm{Ivanova}, \binits{N.}},
\bauthor{\bsnm{Pedersen}, \binits{L.T.}},
\bauthor{\bsnm{Tonboe}, \binits{R.T.}},
\bauthor{\bsnm{Kern}, \binits{S.}},
\bauthor{\bsnm{Heygster}, \binits{G.C.}},
\bauthor{\bsnm{Lavergne}, \binits{T.}},
\bauthor{\bsnm{S{\o}rensen}, \binits{A.M.}},
\bauthor{\bsnm{Saldo}, \binits{R.}},
\bauthor{\bsnm{Dybkj{\ae}r}, \binits{G.}},
\bauthor{\bsnm{Brucker}, \binits{L.}},
\bauthor{\bsnm{Shokr}, \binits{M.}}:
\batitle{Inter-comparison and evaluation of sea ice algorithms: towards further identification of challenges and optimal approach using passive microwave observations}.
\bjtitle{The Cryosphere}
\bvolume{9},
\bfpage{1797}--\blpage{1817}
(\byear{2015})
\end{barticle}
\endbibitem

\bibitem[\protect\citeauthoryear{Chen et~al.}{2023}]{Chen2023CalibrationOU}
\begin{barticle}
\bauthor{\bsnm{Chen}, \binits{X.}},
\bauthor{\bsnm{Valencia}, \binits{R.}},
\bauthor{\bsnm{Soleymani}, \binits{A.}},
\bauthor{\bsnm{Scott}, \binits{K.A.}},
\bauthor{\bsnm{Xu}, \binits{L.}},
\bauthor{\bsnm{Clausi}, \binits{D.A.}}:
\batitle{Calibration of uncertainty in sea ice concentration retrieval with an auxiliary prediction interval estimator}.
\bjtitle{IEEE Geoscience and Remote Sensing Letters}
\bvolume{20},
\bfpage{1}--\blpage{5}
(\byear{2023})
\end{barticle}
\endbibitem

\bibitem[\protect\citeauthoryear{Tschudi et~al.}{2019}]{Tschudi2019EASEGrid}
\begin{botherref}
\oauthor{\bsnm{Tschudi}, \binits{M.}},
\oauthor{\bsnm{Meier}, \binits{W.N.}},
\oauthor{\bsnm{Stewart}, \binits{J.S.}},
\oauthor{\bsnm{Fowler}, \binits{C.}},
\oauthor{\bsnm{Maslanik}, \binits{J.}}:
EASE-Grid Sea Ice Age, Version 4.
NASA National Snow and Ice Data Center Distributed Active Archive Center
(2019).
\doiurl{10.5067/UTAV7490FEPB} .
\url{https://nsidc.org/data/NSIDC-0611/versions/4}
\end{botherref}
\endbibitem

\bibitem[\protect\citeauthoryear{Meier et~al.}{2022}]{Meier2022DMSP}
\begin{botherref}
\oauthor{\bsnm{Meier}, \binits{W.N.}},
\oauthor{\bsnm{Stewart}, \binits{J.S.}},
\oauthor{\bsnm{Wilcox}, \binits{H.}},
\oauthor{\bsnm{Scott}, \binits{D.J.}},
\oauthor{\bsnm{Hardman}, \binits{M.A.}}:
{DMSP SSM/I-SSMIS Daily Polar Gridded Brightness Temperatures, Version 6}.
NASA National Snow and Ice Data Center Distributed Active Archive Center,
Boulder, Colorado USA
(2022)
\end{botherref}
\endbibitem

\bibitem[\protect\citeauthoryear{Lim et~al.}{2023}]{lim2023long}
\begin{bchapter}
\bauthor{\bsnm{Lim}, \binits{S.}},
\bauthor{\bsnm{Park}, \binits{J.}},
\bauthor{\bsnm{Kim}, \binits{S.}},
\bauthor{\bsnm{Wi}, \binits{H.}},
\bauthor{\bsnm{Lim}, \binits{H.}},
\bauthor{\bsnm{Jeon}, \binits{J.}},
\bauthor{\bsnm{Choi}, \binits{J.}},
\bauthor{\bsnm{Park}, \binits{N.}}:
\bctitle{Long-term time series forecasting based on decomposition and neural ordinary differential equations}.
In: \bbtitle{2023 IEEE International Conference on Big Data (BigData)},
pp. \bfpage{748}--\blpage{757}
(\byear{2023}).
\bcomment{IEEE}
\end{bchapter}
\endbibitem

\bibitem[\protect\citeauthoryear{Simonyan and Zisserman}{2014}]{simonyan2014two}
\begin{botherref}
\oauthor{\bsnm{Simonyan}, \binits{K.}},
\oauthor{\bsnm{Zisserman}, \binits{A.}}:
Two-stream convolutional networks for action recognition in videos.
Advances in neural information processing systems
\textbf{27}
(2014)
\end{botherref}
\endbibitem

\bibitem[\protect\citeauthoryear{Ali and Wang}{2022}]{Ali2022MTIceNetA}
\begin{bchapter}
\bauthor{\bsnm{Ali}, \binits{S.}},
\bauthor{\bsnm{Wang}, \binits{J.}}:
\bctitle{Mt-icenet - a spatial and multi-temporal deep learning model for arctic sea ice forecasting}.
In: \bbtitle{2022 IEEE/ACM International Conference on Big Data Computing, Applications and Technologies (BDCAT)},
pp. \bfpage{1}--\blpage{10}
(\byear{2022}).
\bcomment{IEEE}
\end{bchapter}
\endbibitem

\bibitem[\protect\citeauthoryear{Ren and Li}{2023}]{ren2023predicting}
\begin{botherref}
\oauthor{\bsnm{Ren}, \binits{Y.}},
\oauthor{\bsnm{Li}, \binits{X.}}:
Predicting the daily sea ice concentration on a sub-seasonal scale of the pan-arctic during the melting season by a deep learning model.
IEEE Transactions on Geoscience and Remote Sensing
(2023)
\end{botherref}
\endbibitem

\bibitem[\protect\citeauthoryear{Creswell et~al.}{2017}]{creswell2017denoising}
\begin{botherref}
\oauthor{\bsnm{Creswell}, \binits{A.}},
\oauthor{\bsnm{Arulkumaran}, \binits{K.}},
\oauthor{\bsnm{Bharath}, \binits{A.A.}}:
On denoising autoencoders trained to minimise binary cross-entropy.
arXiv preprint arXiv:1708.08487
(2017)
\end{botherref}
\endbibitem

\bibitem[\protect\citeauthoryear{de~Le{\'o}n et~al.}{2022}]{de2022novel}
\begin{barticle}
\bauthor{\bsnm{Le{\'o}n}, \binits{G.}},
\bauthor{\bsnm{Cesbron}, \binits{J.}},
\bauthor{\bsnm{Klein}, \binits{P.}},
\bauthor{\bsnm{Leandri}, \binits{P.}},
\bauthor{\bsnm{Losa}, \binits{M.}}:
\batitle{Novel methodology to recover road surface height maps from illuminated scene through convolutional neural networks}.
\bjtitle{Sensors}
\bvolume{22}(\bissue{17}),
\bfpage{6603}
(\byear{2022})
\end{barticle}
\endbibitem

\bibitem[\protect\citeauthoryear{Loshchilov and Hutter}{2017}]{Loshchilov2017DecoupledWD}
\begin{bchapter}
\bauthor{\bsnm{Loshchilov}, \binits{I.}},
\bauthor{\bsnm{Hutter}, \binits{F.}}:
\bctitle{Decoupled weight decay regularization}.
In: \bbtitle{International Conference on Learning Representations}
(\byear{2017})
\end{bchapter}
\endbibitem

\bibitem[\protect\citeauthoryear{Zeng et~al.}{2022}]{Zeng2022AreTE}
\begin{bchapter}
\bauthor{\bsnm{Zeng}, \binits{A.}},
\bauthor{\bsnm{Chen}, \binits{M.-H.}},
\bauthor{\bsnm{Zhang}, \binits{L.}},
\bauthor{\bsnm{Xu}, \binits{Q.}}:
\bctitle{Are transformers effective for time series forecasting?}
In: \bbtitle{AAAI Conference on Artificial Intelligence}
(\byear{2022})
\end{bchapter}
\endbibitem

\bibitem[\protect\citeauthoryear{Lim et~al.}{2021}]{lim2021temporal}
\begin{barticle}
\bauthor{\bsnm{Lim}, \binits{B.}},
\bauthor{\bsnm{Ar{\i}k}, \binits{S.{\"O}.}},
\bauthor{\bsnm{Loeff}, \binits{N.}},
\bauthor{\bsnm{Pfister}, \binits{T.}}:
\batitle{Temporal fusion transformers for interpretable multi-horizon time series forecasting}.
\bjtitle{International Journal of Forecasting}
\bvolume{37}(\bissue{4}),
\bfpage{1748}--\blpage{1764}
(\byear{2021})
\end{barticle}
\endbibitem

\bibitem[\protect\citeauthoryear{Rakowski and Lippert}{2023}]{rakowski2023dcid}
\begin{bchapter}
\bauthor{\bsnm{Rakowski}, \binits{A.}},
\bauthor{\bsnm{Lippert}, \binits{C.}}:
\bctitle{Dcid: Deep canonical information decomposition}.
In: \bbtitle{Joint European Conference on Machine Learning and Knowledge Discovery in Databases},
pp. \bfpage{20}--\blpage{35}
(\byear{2023}).
\bcomment{Springer}
\end{bchapter}
\endbibitem

\end{thebibliography}

\end{document}